\documentclass{article}

\usepackage{template/iclr2021_conference}

\usepackage[T1]{fontenc}
\usepackage[utf8]{inputenc}
\usepackage[english]{babel}
\usepackage{amsmath}
\usepackage{amsfonts}
\usepackage{bm}
\usepackage{etoolbox}
\usepackage{hyperref}
\usepackage{url}
\usepackage{graphicx}
\usepackage{xcolor}
\usepackage{mathtools}
\usepackage{amssymb}
\usepackage{commath}
\usepackage{stmaryrd}
\usepackage{nicefrac}
\usepackage[inline]{enumitem}
\usepackage[separate-uncertainty=true]{siunitx}
\usepackage{breqn}
\usepackage{microtype}
\usepackage{times}
\usepackage{cases}
\usepackage{import}
\usepackage{booktabs}
\usepackage{transparent}
\usepackage{multirow}
\usepackage{multido}
\usepackage{makecell}
\usepackage{graphics}
\usepackage{natbib}
\usepackage{subfigure}
\usepackage{wrapfig}
\usepackage{comment}
\usepackage[capitalize, noabbrev]{cleveref}
\usepackage{datetime}

\newcommand{\erm}{\mathrm{e}}
\newcommand{\id}{\mathrm{id}}

\DeclarePairedDelimiter\parentheses{(}{)}

\DeclarePairedDelimiter\brackets{[}{]}

\DeclarePairedDelimiter\lrbrackets{\llbracket}{\rrbracket}
\DeclarePairedDelimiter\euclideannorm{\|}{\|}

\DeclarePairedDelimiterX{\midx}[2]{(}{)}{#1\;\delimsize\vert\;#2}
\DeclarePairedDelimiterX{\midbracesx}[2]{\{}{\}}{#1\;\delimsize\vert\;#2}
\DeclarePairedDelimiterX{\parallelx}[2]{(}{)}{#1\;\delimsize\|\;#2}

\def\eqref#1{equation~\ref{#1}}

\def\1{\bm{1}}

\DeclareMathAlphabet{\mathsfit}{\encodingdefault}{\sfdefault}{m}{sl}
\SetMathAlphabet{\mathsfit}{bold}{\encodingdefault}{\sfdefault}{bx}{n}

\def\gB{{\mathcal{B}}}
\def\gC{{\mathcal{C}}}

\def\gL{{\mathcal{L}}}

\def\gP{{\mathcal{P}}}

\def\gS{{\mathcal{S}}}
\def\gT{{\mathcal{T}}}
\def\gU{{\mathcal{U}}}
\def\gV{{\mathcal{V}}}

\def\gX{{\mathcal{X}}}

\def\sN{{\mathbb{N}}}

\def\sR{{\mathbb{R}}}

\newcommand{\E}{\mathbb{E}}

\definecolor{mydarkblue}{rgb}{0,0.08,0.45}
\hypersetup{
    colorlinks=true,
    linkcolor=mydarkblue,
    citecolor=mydarkblue,
    filecolor=mydarkblue,
    urlcolor=mydarkblue,
    pdfview=FitH
}

\AtBeginDocument{}

\title{PDE-Driven Spatiotemporal Disentanglement}

\author{%
    Jérémie Donà,\textsuperscript{\ensuremath{\dagger}}\thanks{Equal contribution.} \ \ Jean-Yves Franceschi,\textsuperscript{\ensuremath{\dagger}}\footnotemark[1] \ \ Sylvain Lamprier\textsuperscript{\ensuremath{\dagger}} \& Patrick Gallinari\textsuperscript{\ensuremath{\dagger\ddagger}} \\
    \textsuperscript{\ensuremath{\dagger}}Sorbonne Université, CNRS, LIP6, F-75005 Paris, France \\
    \textsuperscript{\ensuremath{\ddagger}}Criteo AI Lab, Paris, France \\
    \texttt{firstname.lastname@lip6.fr} \\
}

\iclrfinalcopy 

\begin{document}

\maketitle

\begin{abstract}
    A recent line of work in the machine learning community addresses the problem of predicting high-dimensional spatiotemporal phenomena by leveraging specific tools from the differential equations theory.
    Following this direction, we propose in this article a novel and general paradigm for this task based on a resolution method for partial differential equations: the separation of variables.
    This inspiration allows us to introduce a dynamical interpretation of spatiotemporal disentanglement.
    It induces a principled model based on learning disentangled spatial and temporal representations of a phenomenon to accurately predict future observations.
    We experimentally demonstrate the performance and broad applicability of our method against prior state-of-the-art models on physical and synthetic video datasets.
\end{abstract}

\section{Introduction}
The interest of the machine learning community in physical phenomena has substantially grown for the last few years \citep{Shi2015, Long2018, Greydanus2019}.
In particular, an increasing amount of works studies the challenging problem of modeling the evolution of dynamical systems, with applications in sensible domains like climate or health science, making the understanding of physical phenomena a key challenge in machine learning.
To this end, the community has successfully leveraged the formalism of dynamical systems and their associated differential formulation as powerful tools to specifically design efficient prediction models.
In this work, we aim at studying this prediction problem with a principled and general approach, through the prism of Partial Differential Equations (PDEs), with a focus on learning spatiotemporal disentangled representations.

Prediction via spatiotemporal disentanglement was first studied in video prediction works, in order to separate static and dynamic information \citep{Denton2017} for prediction and interpretability purposes.
Existing models are particularly complex, involving either adversarial losses or variational inference.
Furthermore, their reliance on Recurrent Neural Networks (RNNs) hinders their ability to model spatiotemporal phenomena \citep{Yildiz2019, Ayed2020, Franceschi2020}.
Our proposition addresses these shortcomings with a simplified and improved model by grounding spatiotemporal disentanglement in the PDE formalism.

Spatiotemporal phenomena obey physical laws such as the conservation of energy, that lead to describe the evolution of the system through PDEs.
Practical examples include the conservation of energy for physical systems \citep{Hamilton1835}, or the equation describing constant illumination in a scene \citep{Horn1981} for videos that has had a longstanding impact in computer vision with optical flow methods \citep{Dosovitskiy2015, Finn2016}.
We propose to model the evolution of partially observed spatiotemporal phenomena with unknown dynamics by leveraging a formal method for the analytical resolution of PDEs: the functional separation of variables \citep{Miller1988}.
Our framework formulates spatiotemporal disentanglement for prediction as learning a separable solution, where spatial and dynamic information are represented in separate variables.
Besides offering a novel interpretation of spatiotemporal disentanglement, it confers simplicity and performance compared to existing methods: disentanglement is achieved through the sole combination of a prediction objective and regularization penalties, and the temporal dynamics is defined by a learned Ordinary Differential Equation (ODE).
We experimentally demonstrate the applicability, disentanglement capacity and forecasting performance of the proposed model on various spatiotemporal phenomena involving standard physical processes and synthetic video datasets against prior state-of-the-art models.

\section{Related Work}
\label{sec:related}

Our contribution deals with two main directions of research: spatiotemporal disentanglement and the coupling of neural networks and PDEs.

\paragraph{Spatiotemporal disentanglement.}

Disentangling factors of variations is an essential representation learning problem \citep{Bengio2013}.
Its cardinal formulation for static data has been extensively studied, with state-of-the-art solutions \citep{Locatello2019} being essentially based on Variational Autoencoders \citep[VAEs;][]{Kingma2014, Rezende2014}.
As for sequential data, several disentanglement notions have been formulated, ranging from distinguishing objects in a video \citep{Hsieh2018, Steenkiste2018} to separating and modeling multi-scale dynamics \citep{Hsu2017, Yingzhen2018}.

We focus in this work on the dissociation of the dynamics and visual aspects for spatiotemporal data.
Even in this case, dissociation can take multiple forms.
Examples in the video generation community include decoupling the foreground and background \citep{Vondrick2016}, constructing structured frame representations \citep{Villegas2017b, Minderer2019, Xu2019}, extracting physical dynamics \citep{LeGuen2020}, or latent modeling of dynamics in a state-space manner \citep{Fraccaro2017, Franceschi2020}.
Closer to our work, \cite{Denton2017}, \cite{Villegas2017a} and \cite{Hsieh2018} introduced in their video prediction models explicit latent disentanglement of static and dynamic information obtained using adversarial losses \citep{Goodfellow2014} or VAEs.
Disentanglement has also been introduced in more restrictive models relying on data-specific assumptions \citep{Kosiorek2018, Jaques2020}, and in video generation \citep{Tulyakov2018}.
We aim in this work at grounding and improving spatiotemporal disentanglement with more adapted inductive biases by introducing a paradigm leveraging the functional separation of variables resolution method for PDEs.

\paragraph{Spatiotemporal prediction and PDE-based neural network models.}

An increasing number of works combining neural networks and differential equations for spatiotemporal forecasting have been produced for the last few years.
Some of them show substantial improvements for the prediction of dynamical systems or videos compared to standard RNNs by defining the dynamics using learned ODEs \citep{Rubanova2019, Yildiz2019, Ayed2020, LeGuen2020}, following \cite{Chen2018}, or adapting them to stochastic data \citep{Ryder2018, Li2020, Franceschi2020}.
Most PDE-based spatiotemporal models exploit some prior physical knowledge.
It can induce the structure of the prediction function \citep{Brunton2016, Avila2018} or specific cost functions, thereby improving model performances.
For instance, \cite{Bezenac2018} shape their prediction function with an advection-diffusion mechanism, and \cite{Long2018, Long2019} estimate PDEs and their solutions by learning convolutional filters proven to approximate differential operators.
\cite{Greydanus2019}, \cite{Chen2020} and \cite{Toth2020} introduce non-regression losses by taking advantage of Hamiltonian mechanics \citep{Hamilton1835}, while \cite{Tompson2017} and \cite{Raissi2020} combine physically inspired constraints and structural priors for fluid dynamic prediction.
Our work deepens this literature by establishing a novel link between a resolution method for PDEs and spatiotemporal disentanglement, thereby introducing a data-agnostic model leveraging any static information in observed phenomena.

\section{Background: Separation of Variables}
\label{background}

Solving high-dimensional PDEs is a difficult analytical and numerical problem \citep{Bungartz2004}.
Variable separation aims at simplifying it by decomposing the solution, e.g., as a simple combination of lower-dimensional functions, thus reducing the PDE to simpler differential equations.

\subsection{Simple Case Study}
\label{sec.heat_sep}

Let us introduce this technique through a standard application, with proofs in \cref{sec:heateqdetails}, on the one-dimensional heat diffusion problem \citep{Fourier1822}, consisting in a bar of length $L$, whose temperature at time $t$ and position $x$ is denoted by $u\parentheses*{x, t}$ and satisfies:
\begin{align}
    \label{eq.heat}
    \dpd{u}{t} = c^{2} \dpd[2]{u}{x}, && u\parentheses*{0, t} = u\parentheses*{L, t} = 0, && u\parentheses*{x, 0} = f\parentheses*{x}.
\end{align}
Suppose that a solution $u$ is product-separable, i.e., it can be decomposed as: $u\parentheses*{x, t} = u_{1}\parentheses*{x} \cdot u_{2}\parentheses*{t}$.
Combined with \cref{eq.heat}, it leads to $c^{2} u_{1}''\parentheses*{x} / u_{1}\parentheses*{x} = u_{2}'\parentheses*{t} / u_{2}\parentheses*{t}$.
The left- and right-hand sides of this equation are respectively independent from $t$ and $x$.
Therefore, both sides are constant, and solving both resulting ODEs gives solutions of the form, with $\mu \in \sR$ and $n \in \sN$:
\begin{equation}
    u\parentheses*{x, t} = \mu \sin\parentheses*{n \pi x / L} \times \exp\parentheses*{- \parentheses*{c n \pi / L}^{2} t}.
\end{equation}
The superposition principle and the uniqueness of solutions under smoothness constraints allow then to build the set of solutions of \cref{eq.heat} with linear combinations of separable solutions \citep{LeDret2016}.
Besides this simple example, separation of variables can be more elaborate.

\subsection{Functional Separation of Variables}
The functional separation of variables \citep{Miller1988} generalizes this method.
Let $u$ be a function obeying a given arbitrary PDE.
The functional variable separation method amounts to finding a parameterization $z$, a functional $U$, an entangling function $\xi$, and representations $\phi$ and $\psi$ such that:
\begin{align}
    \label{gen.var_sep}
    z = \xi\parentheses*{\phi\parentheses*{x}, \psi\parentheses*{t}}, && u\parentheses*{x, t} = U\parentheses*{z}.
\end{align}
Trivial choices $\xi = u$ and identity function as $U$, $\phi$ and $\psi$ ensure the validity of this reformulation.
Finding suitable $\phi$, $\psi$, $U$, and $\xi$ with regards to the initial PDE can facilitate its resolution by inducing separate simpler PDEs on $\phi$, $\psi$, and $U$.
For instance, product-separability is retrieved with $U = \exp$.
General results on the existence of separable solutions have been proven \citep{Miller1983}, though their uniqueness depends on the initial conditions and the choice of functional separation \citep{Polyanin2020a}.

Functional separation of variables finds broad applications.
It helps to solve refinements of the heat equation, such as generalizations with an advection term (see \cref{sec:heateq2details}) or with complex diffusion and source terms forming a general transport equation \citep{Jia2008}.
Besides the heat equation, functional separation of PDEs is also applicable in various physics fields like reaction-diffusion with non-linear sources or convection-diffusion phenomena \citep{Polyanin2019, Polyanin2020b}, Hamiltonian physics \citep{Benenti1997}, or even general relativity \citep{Kalnins1992}.

Reparameterizations such as \cref{gen.var_sep} implement a separation of spatial and temporal factors of variations, i.e., spatiotemporal disentanglement.
We introduce in the following a learning framework based on this general method.

\section{Proposed Method}
\label{sec:method}

We propose to model spatiotemporal phenomena using the functional variable separation formalism.
We first describe our notations and then derive a principled model and constraints from this method.

\subsection{Problem Formulation Through Separation of Variables}
\label{sec:formulation}

We consider a distribution $\gP$ of observed spatiotemporal trajectories and corresponding observation samples $v=\parentheses*{v_{t_{0}}, v_{t_{0} + \Delta t}, \ldots, v_{t_{1}}}$, 
with $v_{t} \in \gV \subseteq \sR^{m}$ and $t_{1} = t_{0} + \nu \Delta t$. 
Each sequence $v \sim \gP$ corresponds to an observation of a dynamical phenomenon, assumed to be described by a hidden functional $u_{v}$ (also denoted by $u$ for the sake of simplicity) of space coordinates $x \in \gX \subseteq \sR^{s}$ and time $t \in \sR$ that characterizes the trajectories.
More precisely, $u_{v}$ describes an unobserved continuous dynamics and $v$ corresponds to instantaneous discrete spatial measurements associated to this dynamics. Therefore, we consider that $v_{t}$ results from a time-independent function $\zeta$ of the mapping $u_{v}\parentheses*{\cdot, t}$.
For example, $v$ might consist in temperatures measured at some points of the sea surface, while $u_{v}$ would be the complete ocean circulation model.
In other words, $v$ provides a partial information about $u_{v}$ and is a projection of the full dynamics.
We seek to learn a model which, when conditioned on prior observations, can predict future observations.

To this end, we posit that the state $u$ of each observed trajectory $v$ is driven by a hidden PDE, shared among all trajectories; we discuss this assumption in details in \cref{app:DifferentialEquationsDiscussion}.
Learning such a PDE and its solutions would then allow us to model observed trajectories $v$.
However, directly learning solutions to high-dimensional unknown PDEs is a complex task \citep{Bungartz2004, Sirignano2018}.
We aim in this work at simplifying this resolution.
We propose to do so by relying on the functional separation of variables of \cref{gen.var_sep}, in order to leverage a potential separability of the hidden PDE.
Therefore, analogously to \cref{gen.var_sep}, we propose to formulate the problem as learning observation-constrained $\phi$, $\psi$ and $U$, as well as $\xi$ and $\zeta$, such that:
\begin{align}
    \label{eq.explicit_var_sep}
    z = \xi\parentheses*{\phi\parentheses*{x}, \psi\parentheses*{t}}, && u\parentheses*{x, t} = U\parentheses*{z}, && v_{t} = \zeta\parentheses*{u\parentheses*{\cdot, t}},
\end{align}
with $\phi$ and $\psi$ allowing to disentangle the prediction problem.
In the formalism of the functional separation of variables, this amounts to decomposing the full solution $u$, thereby learning a spatial PDE on $\phi$, a temporal ODE on $\psi$, and a PDE on $U$, as well as their respective solutions.

\subsection{Fundamental Limits and Relaxation}

Directly learning $u$ is, however, a restrictive choice.
Indeed, when formulating PDEs such as in \cref{eq.heat}, spatial coordinates ($x$, $y$, etc.) and time $t$ appear as variables of the solution. 
Yet, unlike in fully observable phenomena studied by \cite{Sirignano2018} and \cite{Raissi2018}, directly accessing theses variables in practice can be costly or infeasible in our partially observed setting.
In other words, the nature and number of these variables are unknown.
For example, the dynamic of the observed sea surface temperature is highly dependent on numerous unobserved variables such as temperature at deeper levels or wind intensity.
Explicitly taking into account these unobserved variables can only be done with prior domain knowledge.
To maintain the generality of the proposed approach, we choose not to make any data-specific assumption on these unknown variables.

We overcome these issues by eliminating the explicit modeling of spatial coordinates by learning dynamic and time-invariant representations accounting respectively for the time-dependent and space-dependent parts of the solution.
Indeed, \cref{eq.explicit_var_sep} induces that these spatial coordinates, hence the explicit resolution of PDEs on $u$ or $U$, can be ignored, as it amounts to learning $\phi$, $\psi$ and $D$ such that:
\begin{equation}
    \label{eq.implicit_var_sep}
    v_{t} = \parentheses*{\zeta \circ U \circ \xi} \parentheses*{\phi\parentheses*{\cdot}, \psi\parentheses*{t}} = D\parentheses*{\phi, \psi\parentheses*{t}}.
\end{equation}
In order to manipulate functionals $\phi$ and $\psi$ in practice, we respectively introduce learnable time-invariant and time-dependent representations of $\phi$ and $\psi$, denoted by $S$ and $T$, such that:
\begin{align}
    \label{eq.var_sep_representation}
    \phi \equiv S \in \gS \subseteq \sR^{d}, && \psi \equiv T \colon t \mapsto T_{t} \in \gT \subseteq \sR^{p},
\end{align}
where the dependence of $\psi \equiv T$ on time $t$ will be modeled using a temporal ODE following the separation of variables, and the function $\phi$, and consequently its spatial PDE, are encoded into a vectorial representation $S$.
Besides their separation of variables basis, the purpose of $S$ and $T$ is to capture spatial and motion information of the data.
For instance, $S$ could encode static information such as objects appearance, while $T$ typically contains motion variables.

\begin{figure}
    \centering
    \includegraphics[width=\textwidth]{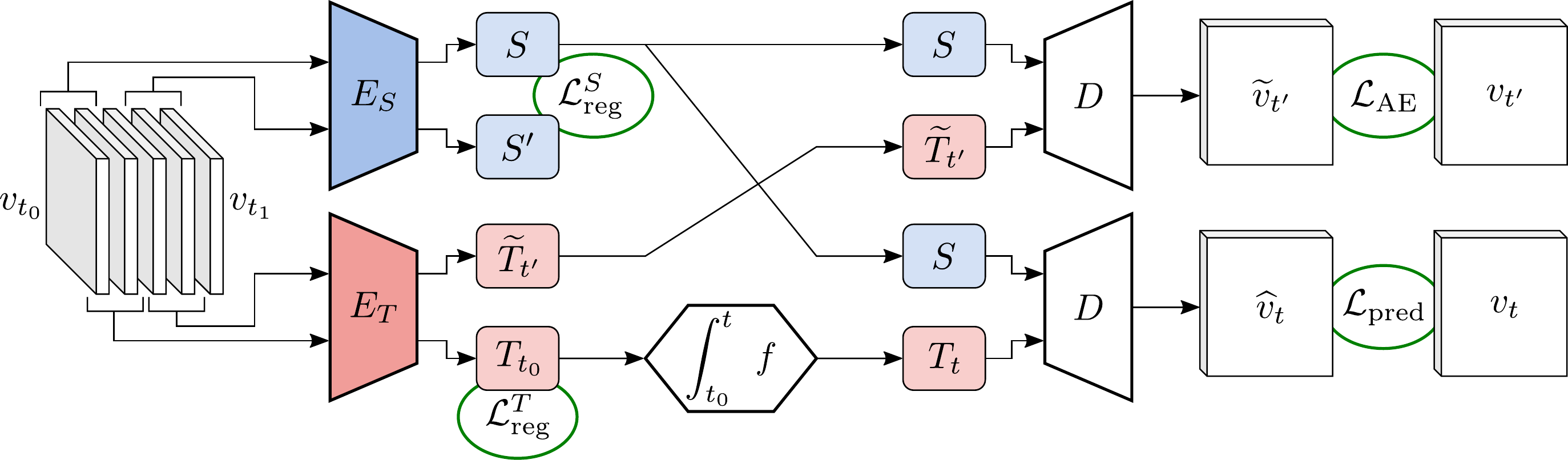}
    \caption{
        \label{fig:algo flow}
        Computational graph of the proposed model.
        $E_{S}$ and $E_{T}$ take contiguous observations as input; time invariance is enforced on $S$; the evolution of $T_{t}$ is modeled with an ODE and is constrained to coincide with $E_{T}$; $T_{t_{0}}$ is regularized; forecasting amounts to decoding from $S$ and $T_{t}$.
    }
\end{figure}

$S$ and $T_{t_{0}}$, because of their dependence on $v$ in \cref{eq.implicit_var_sep,eq.var_sep_representation}, are inferred from an observation history, or conditioning frames, $V_{\tau}\parentheses*{t_{0}}$, where $V_{\tau}\parentheses*{t} = \parentheses*{v_{t}, v_{t + \Delta t}, \ldots, v_{t + \tau \Delta t}}$, using respectively encoder networks $E_{S}$ and $E_{T}$. 
We parameterize $D$ of \cref{eq.implicit_var_sep} as a neural network that acts on both $S$ and $T_{t}$, and outputs the estimated observation $\widehat{v}_{t} = D\parentheses*{S, T_{t}}$.
Unless specified otherwise, $S$ and $T_{t}$ are fed concatenated into $D$, which then learns the parameterization $\xi$ of their combination.

\subsection{Temporal ODE}

The separation of variables allows us to partly reduce the complex task of learning and integrating PDEs to learning and integrating an ODE on $\psi$, which has been extensively studied in the literature, as explained in \cref{sec:related}.
We therefore model the evolution of $T_{t}$, thereby the dynamics of our system, with a first-order ODE:
\begin{align}
    \label{eq.time_ode}
    \dpd{T_{t}}{t} = f\parentheses*{T_{t}} && \Leftrightarrow && T_{t} = T_{t_{0}} + \int_{t_0}^{t}f(T_{t'}) \dif t'
\end{align}
Note that the first-order ODE assumption can be taken without loss of generality since any ODE is equivalent to a higher-dimensional first-order ODE.
Following \cite{Chen2018}, $f$ is implemented by a neural network and \cref{eq.time_ode} is solved with an ODE resolution scheme.
Suppose initial ODE conditions $S$ and $T_{t_{0}}$ have been computed with $E_{S}$ and $E_{T}$.
This leads to the following simple forecasting scheme, enforced by the corresponding regression loss:
\begin{align}
    \label{eq.forecast_loss}
    \widehat{v}_{t} = D\parentheses*{S, T_{t_{0}} + \int_{t_0}^{t}f(T_{t'}) \dif t'}, && %
    \gL_{\mathrm{pred}} = \frac{1}{\nu + 1} \sum_{i = 0}^{\nu} \frac{1}{m} \euclideannorm*{\widehat{v}_{t_{0} + i \Delta t} - v_{t_0 + i \Delta t}}^{2}_{2},
\end{align}
where $\nu + 1$ is the number of observations, and $m$ is the dimension of the observed variables $v$.

\cref{eq.forecast_loss} ensures that the evolution of $T$ is coherent with the observations; we should enforce its consistency with $E_{T}$.
Indeed, the dynamics of $T_{t}$ is modeled by \cref{eq.time_ode}, while only its initial condition $T_{t_{0}}$ is computed with $E_{T}$.
However, there is no guaranty that $T_{t}$, computed via integration, matches $E_{T}\parentheses*{V_{\tau}\parentheses*{t}}$ at any other time $t$, while they should in principle coincide.
We introduce the following autoencoding constraint mitigating their divergence, thereby stabilizing the evolution of $T$:
\begin{align}
    \label{loss.alignment}
    \gL_{\mathrm{AE}} = \frac{1}{m} \euclideannorm*{D\parentheses*{S, E_{T}\parentheses*{V_{\tau}\parentheses*{t_{0} + i \Delta t}}} - v_{t_{0} + i \Delta t}}_{2}^{2}, && \text{with } i \sim \gU\parentheses*{\lrbrackets*{0, \nu - \tau}}. 
\end{align}

\subsection{Spatiotemporal Disentanglement}
\label{sec:disentanglement}

As indicated hereinabove, the spatial PDE on $\phi$ is assumed to be encoded into $S$.
Nonetheless, since $S$ is inferred from an observation history, we need to explicitly enforce its time independence.
In the PDE formalism, this is equivalent to: 
\begin{align}
    \label{s_invar_strict}
    \dpd{E_{S}\parentheses*{V_{\tau}\parentheses*{t}}}{t} = 0 && \Leftrightarrow && \int_{t_{0}}^{t_{1} - \tau \Delta t} \euclideannorm*{\dpd{E_{S}\parentheses*{V_{\tau}\parentheses*{t}}}{t}}_{2}^{2} \dif t = 0.
\end{align}
However, enforcing \cref{s_invar_strict} raises two crucial issues. 
Firstly, in our partially observed setting, there can be variations of observable content, for instance when an object conceals another one.
Therefore, strictly enforcing a null time derivative is not desirable as it prevents $E_{S}$ to extract accessible information that could be obfuscated in the sequence.
Secondly, estimating this derivative in practice in our setting is unfeasible and costly because of the coarse temporal discretization of the data and the computational cost of $E_{S}$; see \cref{sec.derivative} for more details.
We instead introduce a discretized penalty in our minimization objective, discouraging variations of content between two distant time steps, with $d$ being the dimension of $S$:
\begin{equation}
    \label{s_invar}
    \gL^{S}_{\mathrm{reg}} = \frac{1}{d} \euclideannorm*{E_{S}\parentheses*{V_{\tau}\parentheses*{t_{0}}} - E_{S}\parentheses*{V_{\tau}\parentheses*{t_{1} - \tau \Delta t}}}_{2}^{2}.
\end{equation}
It allows us to overcome the previously stated issues by not enforcing a strict invariance of $S$ and removing the need to estimate any time derivative.
Note that this formulation actually originates from \cref{s_invar_strict} using the Cauchy-Schwarz inequality (see \cref{sec.derivative} for a more general derivation).

Abstracting the spatial ODE on $\phi$ from \cref{eq.explicit_var_sep} into a generic representation $S$ leads, without additional constraints, to an underconstrained problem where spatiotemporal disentanglement cannot be guaranteed.
Indeed, $E_{S}$ can be set to zero to satisfy \cref{s_invar} without breaking any prior constraint, because static information is not prevented to be encoded into $T$. Accordingly, information in $S$ and $T$ needs to be segmented.

Thanks to the design of our model, it suffices to ensure that $S$ and $T$ are disentangled at initial time $t_{0}$ for them be to be disentangled at all $t$. Indeed, the mutual information between two variables is preserved by invertible transformations.
\cref{eq.time_ode} is an ODE and $f$, as a neural network, is Lipschitz-continuous, so the ODE flow $T_{t} \mapsto T_{t'}$ is invertible.
Therefore, disentanglement between $S$ and $T_{t}$, characterized by a low mutual information between both variables, is preserved through time; see \cref{app.disentanglement} for a detailed discussion.
We thus only constrain the information quantity in $T_{t_{0}}$ by using a Gaussian prior to encourage it to exclusively contain necessary dynamic information:
\begin{equation}
    \label{loss.disentanglement}
    \gL_{\mathrm{reg}}^{T} = \frac{1}{p} \euclideannorm*{T_{t_{0}}}_{2}^{2} = \frac{1}{p} \euclideannorm*{E_{T}\parentheses*{V_{\tau}\parentheses*{t_{0}}}}_{2}^{2}.
\end{equation}

\subsection{Loss Function}

The minimized loss is a linear combination of \cref{s_invar,eq.forecast_loss,loss.disentanglement,loss.alignment}:
\begin{equation}
    \begin{multlined}
        \mathcal{L}\parentheses{v} = \E_{v \sim \gP} \brackets*{\lambda_{\mathrm{pred}} \gL_{\mathrm{pred}} + \lambda_{\mathrm{AE}} \cdot \gL_{\mathrm{AE}} + \lambda_{\mathrm{reg}}^{S} \cdot \gL_{\mathrm{reg}}^{S} + \lambda_{\mathrm{reg}}^{T} \cdot \gL_{\mathrm{reg}}^{T}},
    \end{multlined}
\end{equation}
as illustrated in \cref{fig:algo flow}.
In the following, we conventionally set $\Delta t = 1$.
Note that the presented approach could be generalized to irregularly sampled observation times thanks to the dedicated literature \citep{Rubanova2019}, but this is out of the scope of this paper.

\section{Experiments}

\begin{wrapfigure}{R}{0.41\textwidth}
    \centering
    \vspace{-1em}
    \includegraphics[width=0.41\textwidth]{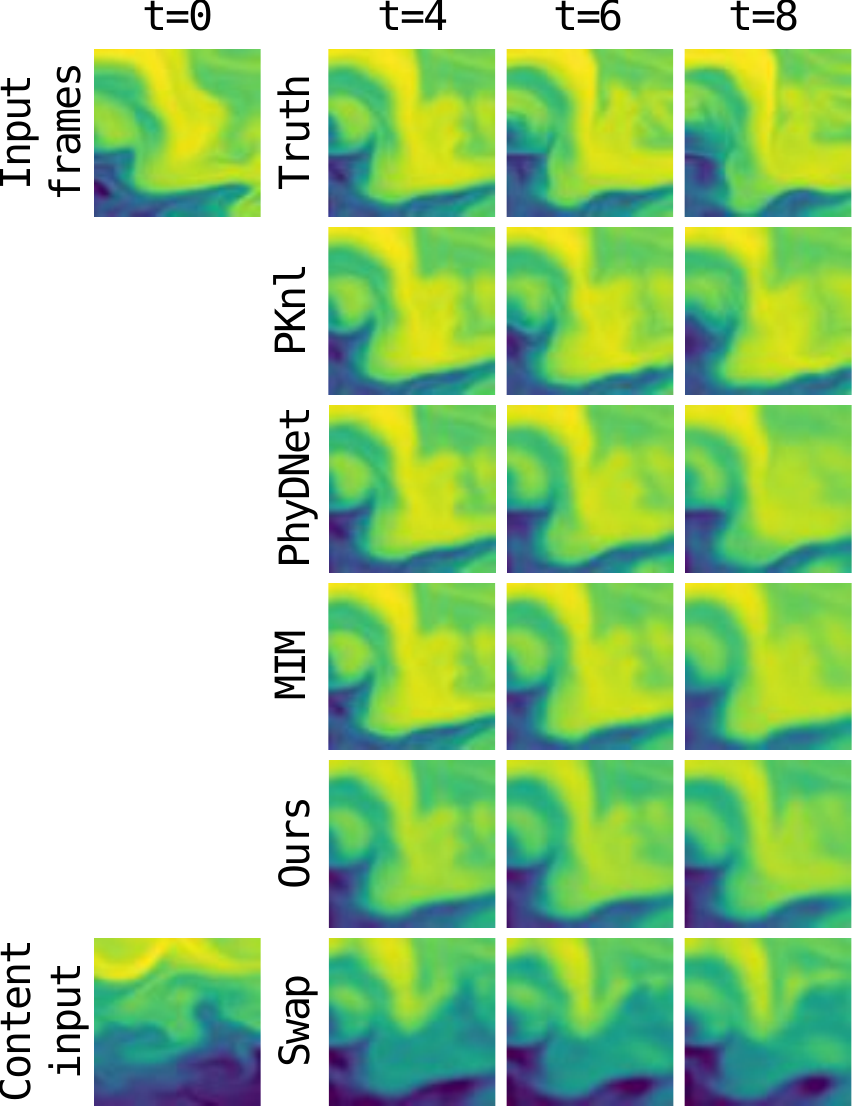}
    \caption{
        \label{fig:sst}
        Example of predictions of compared models on SST.
        Content swap preserves the location of extreme temperature regions which determine the movement while modifying the magnitude of all regions, especially in temperate areas.
    }
    \vspace{-2.3em}
\end{wrapfigure}

We study in this section the experimental results of our model on various spatiotemporal phenomena with physical, synthetic video and real-world datasets, which are briefly presented in this section and in more details in \cref{sec:datasets}.
We demonstrate the relevance of our model with ablation studies and its performance by comparing it with more complex state-of-the-art models.
Performances are assessed thanks to standard metrics \citep{Denton2018, LeGuen2020} Mean Squared Error (MSE, lower is better) or its alternative Peak Signal-to-Noise Ratio (PSNR, higher is better), and Structured Similarity (SSIM, higher is better).
We refer to \cref{sec:additional} for more experiments and prediction examples, to \cref{sec:training} for training information and to the supplementary material for the corresponding code\footnote{Our source code is also publicly released at the following URL: \url{https://github.com/JeremDona/spatiotemporal_variable_separation}.} and datasets.

\subsection{Physical Datasets: Wave Equation and Sea Surface Temperature}

We first investigate two synthetic dynamical systems and a real-world dataset in order to show the advantage of PDE-driven spatiotemporal disentanglement for forecasting physical phenomena.
To analyze our model, we first lean on the wave equation, occurring for example in acoustic or electromagnetism, with source term like \cite{Saha2020}, to produce the WaveEq dataset consisting in $64 \times 64$ normalized images of the phenomenon.
We additionally build the WaveEq-100 dataset by extracting $100$ pixels, chosen uniformly at random and shared among sequences, from WaveEq frames; this experimental setting can be thought of as measurements from sensors partially observing the phenomenon.
We also test and compare our model on the real-world dataset SST, derived from the data assimilation engine NEMO \citep{Madec2008} and introduced by \cite{Bezenac2018}, consisting in $64 \times 64$ frames showing the evolution of the sea surface temperature.
Modeling its evolution is particularly challenging as its dynamic is highly non-linear, chaotic, and involves several unobserved quantities (e.g., forcing terms).

\begin{table}
    \caption{
        \label{tab:phy-res}
        Forecasting performance on WaveEq-100, WaveEq and SST of compared models with respect to indicated prediction horizons.
        Bold scores indicate the best performing method.
    }
    \sisetup{detect-weight, table-align-uncertainty=true, table-number-alignment=center, mode=text}
    \renewrobustcmd{\bfseries}{\fontseries{b}\selectfont}
    \renewrobustcmd{\boldmath}{}
    \centering
    \vspace{\baselineskip}
    \begin{tabular}{lS[table-format=1.2, table-figures-exponent=2]S[table-format=1.2, table-figures-exponent=2]S[table-format=1.2]S[table-format=1.2]S[table-format=1.4]S[table-format=1.4]}
        \toprule
        \multirow{3}{*}[-\dimexpr \aboverulesep + \belowrulesep + \cmidrulewidth]{Models} & {WaveEq-100} & {WaveEq} & \multicolumn{4}{c}{SST} \\
        \cmidrule(lr){2-2} \cmidrule(lr){3-3} \cmidrule(lr){4-7}
        & \multicolumn{2}{c}{MSE} & \multicolumn{2}{c}{MSE} & \multicolumn{2}{c}{SSIM} \\
        \cmidrule(lr){2-3} \cmidrule(lr){4-5} \cmidrule(lr){6-7}
        & {$t + 40$} & {$t + 40$} & {$t + 6$} & {$t + 10$} & {$t + 6$} & {$t + 10$} \\
        \midrule
        PKnl & {\textemdash} & {\textemdash} & 1.28 & 2.03 & 0.6686 & 0.5844 \\
        PhyDNet & {\textemdash} & {\textemdash} & 1.27 & 1.91 & 0.5782 & 0.4645 \\
        SVG & {\textemdash} & {\textemdash} & 1.51 & 2.06 & 0.6259 & 0.5595 \\
        MIM & {\textemdash} & {\textemdash} & 0.91 & 1.45 & 0.7406 & 0.6525 \\
        \midrule
        Ours & \bfseries 4.33e-5 & \bfseries 1.44e-4 & \bfseries 0.86 & \bfseries 1.43 & \bfseries 0.7466 & \bfseries 0.6577 \\
        Ours (without $S$) & 1.33e-4 & 5.09e-4 & 0.95 & 1.50 & 0.7204 & 0.6446 \\
        \bottomrule
    \end{tabular}
\end{table}

We compare our model on these three datasets to its alternative version with $S$ removed and integrated into $T$, thus also removing $\gL^{S}_{\mathrm{reg}}$ and $\gL^{T}_{\mathrm{reg}}$.
We also include the state-of-the-art PhyDNet \citep{LeGuen2020}, MIM \citep{Wang2019}, SVG \citep{Denton2018} and SST-specific PKnl \citep{Bezenac2018} in the comparison on SST; only PhyDNet and PKnl were originally tested on this dataset by their authors.
Results are compiled in \cref{tab:phy-res} and an example of prediction is depicted in \cref{fig:sst}.

On these three datasets, our model produces more accurate long-term predictions with $S$ than without it.
This indicates that learning an invariant component facilitates training and improves generalization.
The influence of $S$ can be observed by replacing the $S$ of a sequence by another one extracted from another sequence, changing the aspect of the prediction, as shown in \cref{fig:sst} (swap row).
We provide in \cref{sec:additional} further samples showing the influence of $S$ in the prediction.
Even though there is no evidence of intrinsic separability in SST, our trained algorithm takes advantage of its time-invariant component.
Indeed, our model outperforms PKnl despite the data-specific structure of the latter, the stochastic SVG and the high-capacities PhyDNet and MIM model, whereas removing its static component suppresses our advantage.

We highlight that MIM is a computationally-heavy model that manipulates in an autoregressive way $64$ times larger latent states than ours, hence its better reconstruction ability at the first time step.
However, its sharpness and movement gradually vanish, explaining its lower performance than ours.
We refer to \cref{sst-model} for additional discussion on the application of our method and its performance on SST.

\begin{table}
    \caption{
        \label{tab:moving-mnist}
        Prediction and content swap scores of all compared models on Moving MNIST.
        Bold scores indicate the best performing method.
    }
    \sisetup{detect-weight, table-align-uncertainty=true, table-number-alignment=center, mode=text}
    \renewrobustcmd{\bfseries}{\fontseries{b}\selectfont}
    \renewrobustcmd{\boldmath}{}
    \centering
    \vspace{\baselineskip}
    \begin{tabular}{lS[table-format=2.2]S[table-format=1.4]S[table-format=2.2]S[table-format=1.4]S[table-format=2.2]S[table-format=1.4]S[table-format=2.2]S[table-format=1.4]}
        \toprule
        \multirow{2}{*}[-0.5\dimexpr \aboverulesep + \belowrulesep + \cmidrulewidth]{Models} & \multicolumn{2}{c}{Pred. ($t + 10$)} & \multicolumn{2}{c}{Pred. ($t + 95$)} & \multicolumn{2}{c}{Swap ($t + 10$)} & \multicolumn{2}{c}{Swap ($t + 95$)} \\
        \cmidrule(lr){2-3} \cmidrule(lr){4-5} \cmidrule(lr){6-7} \cmidrule(lr){8-9}
        & {PSNR} & {SSIM} & {PSNR} & {SSIM} & {PSNR} & {SSIM} & {PSNR} & SSIM \\
        \midrule
        SVG & 18.18 & 0.8329 & 12.85 & 0.6185 & {\textemdash} & {\textemdash} & {\textemdash} & {\textemdash} \\
        MIM & \bfseries 24.16 & 0.9113 & 16.50 & 0.6529 & {\textemdash} & {\textemdash} & {\textemdash} & {\textemdash} \\
        DrNet & 14.94 & 0.6596 & 12.91 & 0.5379 & 14.12 & 0.6206 & 12.80 & 0.5306 \\
        DDPAE & 21.17 & 0.8814 & 13.56 & 0.6446 & \bfseries 18.44 & 0.8256 & 13.25 & 0.6378 \\
        PhyDNet & 23.12 & \bfseries 0.9128 & 16.46 & 0.3878 & 12.04 & 0.5572 & 13.49 & 0.2839 \\
        Ours & 21.70 & 0.9088 & \bfseries 17.50 & \bfseries 0.7990 & 18.42 & \bfseries 0.8368 & \bfseries 16.50 & \bfseries 0.7713 \\
        \bottomrule
    \end{tabular}
\end{table}

\subsection{A Synthetic Video Dataset: Moving MNIST}
\label{sec:mnist-exp}

We also assess the prediction and disentanglement performance of our model on the Moving MNIST dataset \citep{Srivastava2015} involving MNIST digits \citep{LeCun1998} bouncing over frame borders.
This dataset is particularly challenging in the literature for long-term prediction tasks.
We compare our model to competitive baselines: the non-disentangled SVG \citep{Denton2018} and MIM \citep{Wang2019}, as well as forecasting models with spatiotemporal disentanglement ablities DrNet \citep{Denton2017}, DDPAE \citep{Hsieh2018} and PhyDNet.
We highlight that all these models leverage powerful machine learning tools such as adversarial losses, VAEs and high-capacity temporal architectures, whereas ours is solely trained using regression penalties and small-size latent representations.
We perform as well a full ablation study of our model to confirm the relevance of the introduced method.

\begin{figure}
    \centering
    \begin{minipage}{0.56\textwidth}
        \centering
        \includegraphics[width=\linewidth]{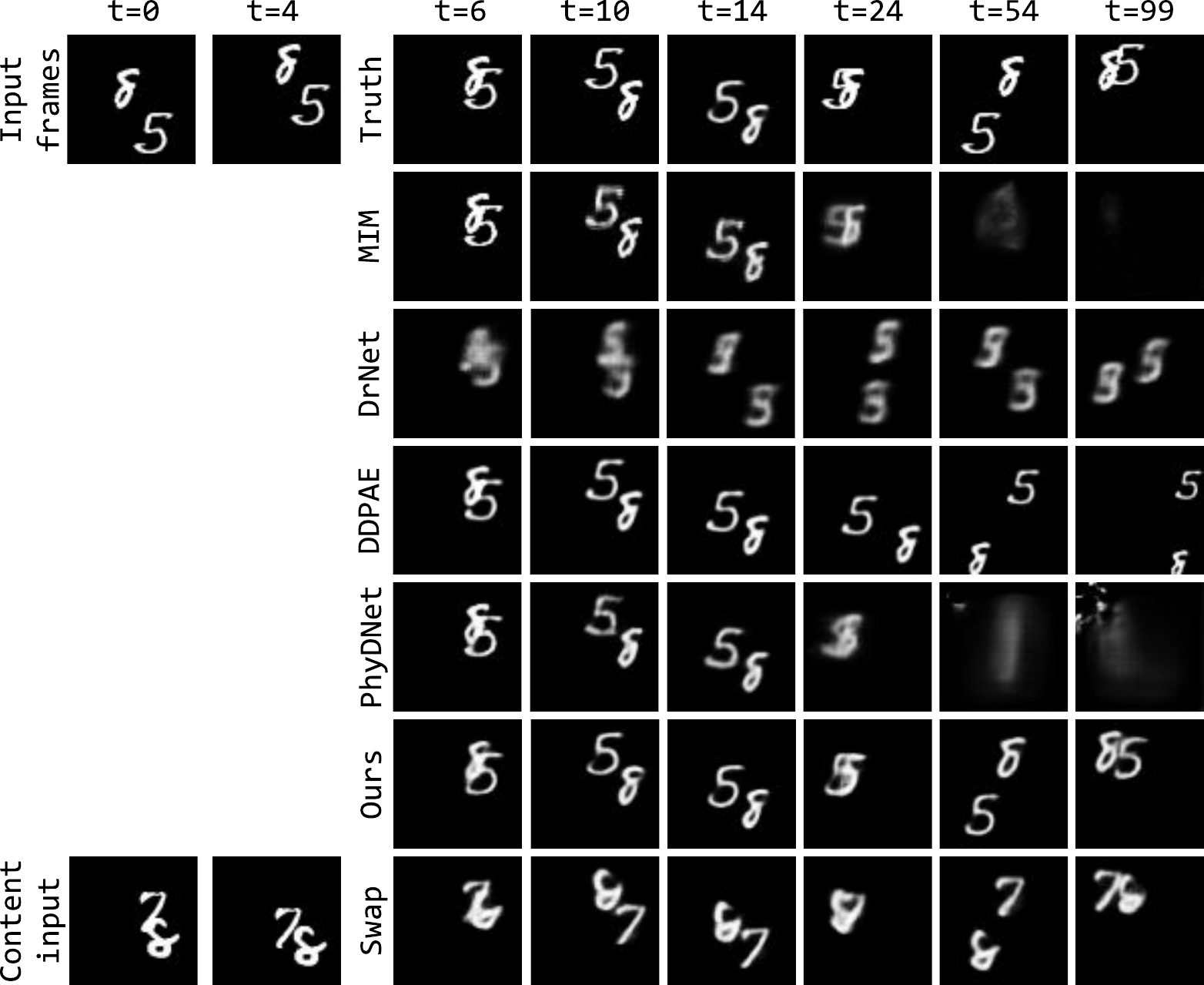}
        \caption{\label{fig:mnist}Predictions of compared models on Moving MNIST, and content swap experiment for our model.}
    \end{minipage}%
    \hfill
    \begin{minipage}{0.422\textwidth}
        \centering
        \includegraphics[width=\linewidth]{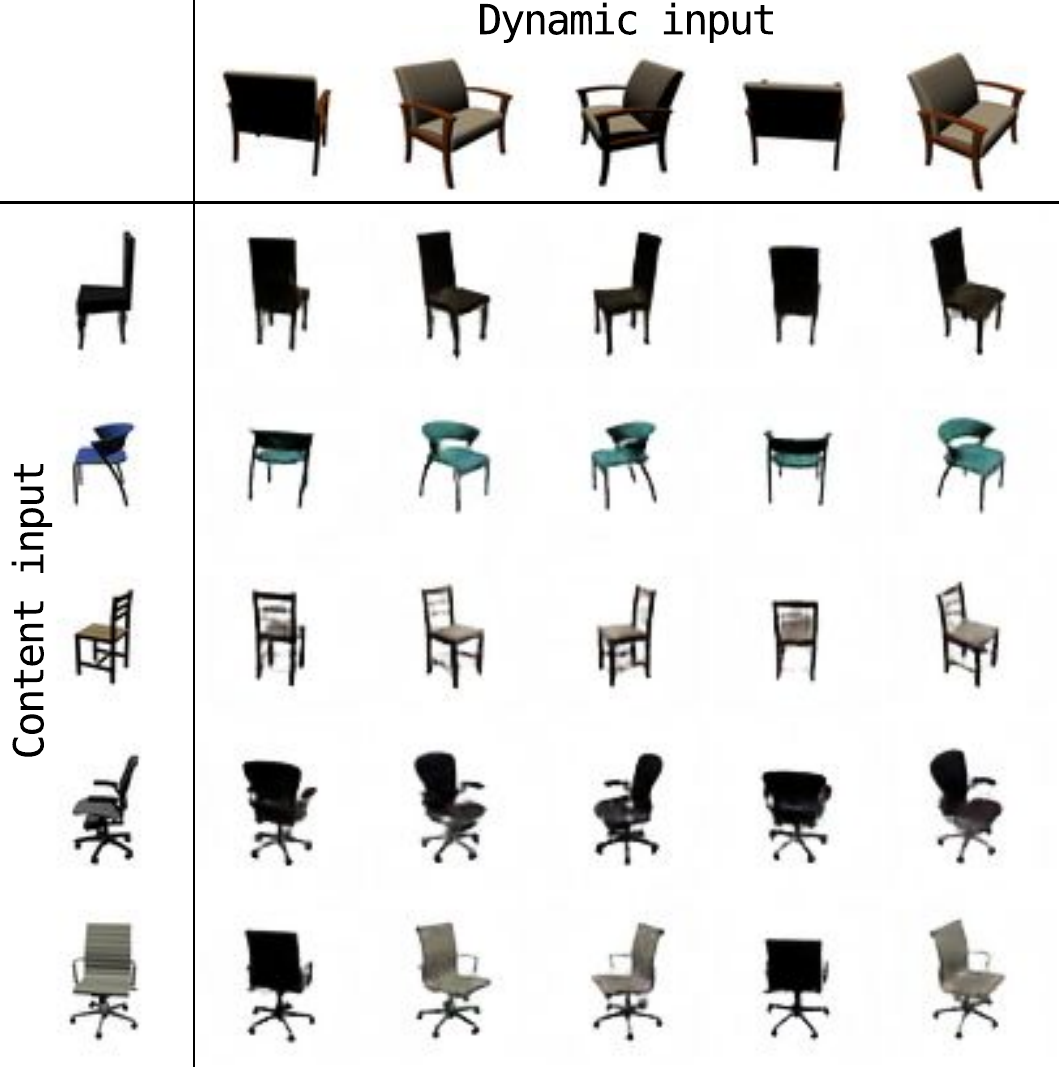}
        \caption{\label{fig:chairs}Fusion of content (first column) and dynamic (first row) variables in our model on 3D Warehouse Chairs.}
    \end{minipage}
\end{figure}

Results reported in \cref{tab:moving-mnist} and illustrated in \cref{fig:mnist} correspond to two tasks: prediction and disentanglement, at both short and long-term horizons.
Disentanglement is evaluated via content swapping, which consists in replacing the content representation of a sequence by the one of another sequence, which should result for a perfectly disentangled model in swapping digits of both sequences.
This is done by taking advantage of the synthetic nature of this dataset that allows us to implement the ground truth content swap and compare it to the generated swaps of the model.

Reported results show the advantage of our model against all baselines.
Long-term prediction challenges them as their performance and predictions collapse in the long run.
This shows that the baselines, including high-capacity models MIM and PhyDNet that leverage powerful ConvLTSMs \citep{Shi2015}, have difficulties separating content and motion.
Indeed, a model separating correctly content and motion should maintain digits appearance even when it miscalculates their trajectories, like DDPAE which alters only marginally the digits in \cref{fig:mnist}.
In contrast, ours manages to produce consistent samples even at $t + 95$, making it reach state-of-the-art performance.
Moreover, we significantly outperform all baselines in the content swap experiment, showing the clear advantage of the proposed PDE-inspired simple model for spatiotemporally disentangled prediction.

Ablation studies developed in \cref{tab:moving-mnist-ablation} confirm that this advantage is due to the constraints motivated by the separation of variables.
Indeed, the model without $S$ fails at long-term forecasting, and removing any non-prediction penalty of the training loss substantially harms performances.
In particular, the invariance loss on the static component and the regularization of initial condition $T_{t_{0}}$ are essential, as their absence hinders both prediction and disentanglement.
The auto-encoding constraint makes predictions more stable, allowing accurate long-term forecasting and disentanglement.
This ablation study also confirms the necessity to constrain the $\ell_{2}$ norm of the dynamic variable (see \cref{loss.disentanglement}) for the model to disentangle.
Comparisons of \cref{tab:moving-mnist} actually show that enforcing this loss on the first time step only is sufficient to ensure state-of-the-art disentanglement, as advocated in \cref{sec:disentanglement}.

Finally, we assess whether the temporal ODE of \cref{eq.time_ode} induced by the separation of variables is advantageous by replacing the dynamic model with a standard GRU RNN \citep{Cho2014}.
Results reported in \cref{tab:moving-mnist-ablation} show substantially better prediction and disentanglement performance for the original model grounded on the separation of variables, indicating the relevance of our approach.

\subsection{A Multi-View Dataset: 3D Warehouse Chairs}

We perform an additional disentanglement experiment on the 3D Warehouse Chairs dataset introduced by \cite{Aubry2014}.
This dataset contains \num{1393} three-dimensional models of chairs seen under various angles.
Since all chairs are observed from the same set of angles, this constitutes a multi-view dataset enabling quantitative disentanglement experiments.
We create sequences from this dataset for our model by assembling adjacent views of each chair to simulate its rotation from right to left.
We then evaluate the disentanglement properties of our model with the same content swap experiments as for Moving MNIST.
It is similar to one of \cite{Denton2017}'s experiments who qualitatively tested their model on a similar, but smaller, multi-view chairs dataset.
We achieve 18.70 PSNR and 0.7746 SSIM on this task, outperforming DrNet which only reaches 16.35 PSNR and 0.6992 SSIM.
This is corroborated by qualitative experiments in \cref{fig:chairs,fig:ChairsOursDenton}. 
We highlight that the encoder and decoder architectures of both competitors are identical, validating our PDE-grounded framework for spatiotemporal disentanglement of complex three-dimensional shapes.

\begin{table}
    \caption{
        \label{tab:taxibj}
        Prediction MSE ($\times 100 \times 32 \times 32 \times 2$) of compared models on TaxiBJ, with best MSE highlighted in bold.
    }
    \sisetup{detect-weight, table-align-uncertainty=true, table-number-alignment=center, mode=text}
    \renewrobustcmd{\bfseries}{\fontseries{b}\selectfont}
    \renewrobustcmd{\boldmath}{}
    \centering
    \vspace{\baselineskip}
    \begin{tabular}{S[table-format=2.1]S[table-format=2.1]S[table-format=2.1]S[table-format=2.1]S[table-format=2.1]S[table-format=2.1]S[table-format=2.1]S[table-format=2.1]}
        \toprule
        {Ours} & {Ours (without $S$)} & {PhyDNet} & {MIM} & {E3D} & {C. LSTM} & {PredRNN} & {ConvLSTM} \\
        \midrule
        \bfseries 39.5 & 43.7 & 41.9 & 42.9 & 43.2 & 44.8 & 46.4 & 48.5 \\
        \bottomrule
    \end{tabular}
\end{table}

\subsection{A Crowd Flow Dataset: TaxiBJ}

\begin{wrapfigure}{R}{0.43\textwidth}
    \centering
    \vspace{-1em}
    \includegraphics[width=0.43\textwidth]{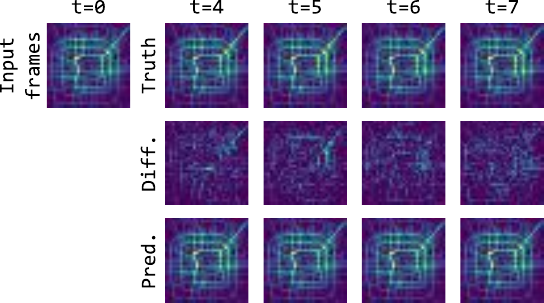}
    \caption{
        \label{fig:taxibj}
        Example of ground truth and prediction of our model on TaxiBJ.
        The middle row shows the scaled difference between our predictions and the ground truth.
    }
    \vspace{-1em}
\end{wrapfigure}

We finally study the performance of our spatiotemporal model on the real-world TaxiBJ dataset \citep{Zhang2017}, consisting in taxi traffic flow in Beijing monitored on a $32 \times 32$ grid with an observation every thirty minutes.
It is highly structured as the flows are dependent on the infrastructures of the city, and complex since methods have to account for non-local dependencies and model subtle changes in the evolution of the flows.
It is a standard benchmark in the spatiotemporal prediction community \citep{Wang2019, LeGuen2020}.

We compare our model in \cref{tab:taxibj} against PhyDNet and MIM, as well as powerful baselines E3D-LSTM \citep[E3D,][]{Wang2019b}, Causal LSTM \citep[C. LSTM,][]{Wang2018}, PredRNN \citep{Wang2017} and ConvLTSM \citep{Shi2015}, using results reported by \cite{Wang2019} and \cite{LeGuen2020}.
An example of prediction is given in \cref{fig:taxibj}.
We observe that we significantly overcome the state of the art on this complex spatiotemporal dataset.
This improvement is notably driven by the disentanglement abilities of our model, as we observe in \cref{tab:taxibj} that the alternative version of our model without $S$ achieves results comparable to E3D and worse than PhyDNet and MIM.

\section{Conclusion}

We introduce a novel method for spatiotemporal prediction inspired by the separation of variables PDE resolution technique that induces time invariance and regression penalties only.
These constraints ensure the separation of spatial and temporal information.
We experimentally demonstrate the benefits of the proposed model, which outperforms prior state-of-the-art methods on physical and synthetic video datasets.
We believe that this work, by providing a dynamical interpretation of spatiotemporal disentanglement, could serve as the basis of more complex models further leveraging the PDE formalism.
Another direction for future work could be extending the model with more involved tools such as VAEs to improve its performance, or adapt it to the prediction of natural stochastic videos \citep{Denton2018}.

\subsubsection*{Acknowledgments}

We would like to thank all members of the MLIA team from the LIP6 laboratory of Sorbonne Université for helpful discussions and comments, as well as Vincent Le Guen for his help to reproduce PhyDNet results and process the TaxiBJ dataset.

We acknowledge financial support from the LOCUST ANR project (ANR-15-CE23-0027) and the European Union’s Horizon 2020 research and innovation programme under grant agreement 825619 (AI4EU).
This study has been conducted using E.U. Copernicus Marine Service Information.
This work was granted access to the HPC resources of IDRIS under allocations 2020-AD011011360 and 2021-AD011011360R1 made by GENCI (Grand Equipement National de Calcul Intensif).
Patrick Gallinari is additionally funded by the 2019 ANR AI Chairs program via the DL4CLIM project.

\bibliography{references.bib}

\begin{thebibliography}{88}
\providecommand{\natexlab}[1]{#1}
\providecommand{\url}[1]{\texttt{#1}}
\expandafter\ifx\csname urlstyle\endcsname\relax
  \providecommand{\doi}[1]{doi: #1}\else
  \providecommand{\doi}{doi: \begingroup \urlstyle{rm}\Url}\fi

\bibitem[Achille \& Soatto(2018)Achille and Soatto]{Achille2018}
Alessandro Achille and Stefano Soatto.
\newblock Emergence of invariance and disentanglement in deep representations.
\newblock \emph{Journal of Machine Learning Research}, 19\penalty0
  (50):\penalty0 1--34, 2018.

\bibitem[Aubry et~al.(2014)Aubry, Maturana, Efros, Russell, and
  Sivic]{Aubry2014}
Mathieu Aubry, Daniel Maturana, Alexei~A. Efros, Bryan~C. Russell, and Josef
  Sivic.
\newblock Seeing {3D} chairs: Exemplar part-based {2D-3D} alignment using a
  large dataset of {CAD} models.
\newblock In \emph{Proceedings of the IEEE Conference on Computer Vision and
  Pattern Recognition (CVPR)}, pp.\  3762--3769, June 2014.

\bibitem[Ayed et~al.(2020)Ayed, de~Bézenac, Pajot, and Gallinari]{Ayed2020}
Ibrahim Ayed, Emmanuel de~Bézenac, Arthur Pajot, and Patrick Gallinari.
\newblock Learning the spatio-temporal dynamics of physical processes from
  partial observations.
\newblock In \emph{ICASSP 2020 - 2020 IEEE International Conference on
  Acoustics, Speech and Signal Processing (ICASSP)}, pp.\  3232--3236, 2020.

\bibitem[Behrmann et~al.(2019)Behrmann, Grathwohl, Chen, Duvenaud, and
  Jacobsen]{Behrmann2019}
Jens Behrmann, Will Grathwohl, Ricky T.~Q. Chen, David Duvenaud, and
  Jörn-Henrik Jacobsen.
\newblock Invertible residual networks.
\newblock In Kamalika Chaudhuri and Ruslan Salakhutdinov (eds.),
  \emph{Proceedings of the 36th International Conference on Machine Learning},
  volume~97 of \emph{Proceedings of Machine Learning Research}, pp.\  573--582,
  Long Beach, California, USA, June 2019. PMLR.

\bibitem[Benenti(1997)]{Benenti1997}
Sergio Benenti.
\newblock Intrinsic characterization of the variable separation in the
  {Hamilton-Jacobi} equation.
\newblock \emph{Journal of Mathematical Physics}, 38\penalty0 (12):\penalty0
  6578--6602, 1997.

\bibitem[Bengio et~al.(2013)Bengio, Courville, and Vincent]{Bengio2013}
Yoshua Bengio, Aaron Courville, and Pascal Vincent.
\newblock Representation learning: A review and new perspectives.
\newblock \emph{IEEE Transactions on Pattern Analysis and Machine
  Intelligence}, 35\penalty0 (8):\penalty0 1798--1828, August 2013.

\bibitem[Brunton et~al.(2016)Brunton, Proctor, and Kutz]{Brunton2016}
Steven~L. Brunton, Joshua~L. Proctor, and J.~Nathan Kutz.
\newblock Discovering governing equations from data by sparse identification of
  nonlinear dynamical systems.
\newblock \emph{Proceedings of the National Academy of Sciences}, 113\penalty0
  (15):\penalty0 3932--3937, 2016.

\bibitem[Bungartz \& Griebel(2004)Bungartz and Griebel]{Bungartz2004}
Hans-Joachim Bungartz and Michael Griebel.
\newblock Sparse grids.
\newblock \emph{Acta Numerica}, 13:\penalty0 147--269, 2004.

\bibitem[Chen et~al.(2018)Chen, Rubanova, Bettencourt, and Duvenaud]{Chen2018}
Ricky T.~Q. Chen, Yulia Rubanova, Jesse Bettencourt, and David Duvenaud.
\newblock Neural ordinary differential equations.
\newblock In Samy Bengio, Hanna Wallach, Hugo Larochelle, Kristen Grauman,
  Nicolò Cesa-Bianchi, and Roman Garnett (eds.), \emph{Advances in Neural
  Information Processing Systems 31}, pp.\  6571--6583. Curran Associates,
  Inc., 2018.

\bibitem[Chen et~al.(2016)Chen, Duan, Houthooft, Schulman, Sutskever, and
  Abbeel]{Chen2016}
Xi~Chen, Yan Duan, Rein Houthooft, John Schulman, Ilya Sutskever, and Pieter
  Abbeel.
\newblock Info{GAN}: Interpretable representation learning by information
  maximizing generative adversarial nets.
\newblock In Daniel~D. Lee, Masashi Sugiyama, Ulrike von Luxburg, Isabelle
  Guyon, and Roman Garnett (eds.), \emph{Advances in Neural Information
  Processing Systems 29}, pp.\  2172--2180. Curran Associates, Inc., 2016.

\bibitem[Chen et~al.(2020)Chen, Zhang, Arjovsky, and Bottou]{Chen2020}
Zhengdao Chen, Jianyu Zhang, Martin Arjovsky, and Léon Bottou.
\newblock Symplectic recurrent neural networks.
\newblock In \emph{International Conference on Learning Representations}, 2020.

\bibitem[Cho et~al.(2014)Cho, van Merriënboer, Gulcehre, Bahdanau, Bougares,
  Schwenk, and Bengio]{Cho2014}
Kyunghyun Cho, Bart van Merriënboer, Caglar Gulcehre, Dzmitry Bahdanau, Fethi
  Bougares, Holger Schwenk, and Yoshua Bengio.
\newblock Learning phrase representations using {RNN} encoder-decoder for
  statistical machine translation.
\newblock In \emph{Proceedings of the 2014 Conference on Empirical Methods in
  Natural Language Processing ({EMNLP})}, pp.\  1724--1734, Doha, Qatar,
  October 2014. Association for Computational Linguistics.

\bibitem[de~Avila Belbute-Peres et~al.(2018)de~Avila Belbute-Peres, Smith,
  Allen, Tenenbaum, and Kolter]{Avila2018}
Filipe de~Avila Belbute-Peres, Kevin~A. Smith, Kelsey~R. Allen, Joshua~B.
  Tenenbaum, and J.~Zico Kolter.
\newblock End-to-end differentiable physics for learning and control.
\newblock In Samy Bengio, Hanna Wallach, Hugo Larochelle, Kristen Grauman,
  Nicolò Cesa-Bianchi, and Roman Garnett (eds.), \emph{Advances in Neural
  Information Processing Systems 31}, pp.\  7178--7189. Curran Associates,
  Inc., 2018.

\bibitem[de~Bézenac et~al.(2018)de~Bézenac, Pajot, and
  Gallinari]{Bezenac2018}
Emmanuel de~Bézenac, Arthur Pajot, and Patrick Gallinari.
\newblock Deep learning for physical processes: Incorporating prior scientific
  knowledge.
\newblock In \emph{International Conference on Learning Representations}, 2018.

\bibitem[Denton \& Birodkar(2017)Denton and Birodkar]{Denton2017}
Emily Denton and Vighnesh Birodkar.
\newblock Unsupervised learning of disentangled representations from video.
\newblock In Isabelle Guyon, Ulrike von Luxburg, Samy Bengio, Hanna Wallach,
  Rob Fergus, S.~V.~N. Vishwanathan, and Roman Garnett (eds.), \emph{Advances
  in Neural Information Processing Systems 30}, pp.\  4414--4423. Curran
  Associates, Inc., 2017.

\bibitem[Denton \& Fergus(2018)Denton and Fergus]{Denton2018}
Emily Denton and Rob Fergus.
\newblock Stochastic video generation with a learned prior.
\newblock In Jennifer Dy and Andreas Krause (eds.), \emph{Proceedings of the
  35th International Conference on Machine Learning}, volume~80 of
  \emph{Proceedings of Machine Learning Research}, pp.\  1174--1183,
  Stockholmsmässan, Stockholm, Sweden, July 2018. PMLR.

\bibitem[Dosovitskiy et~al.(2015)Dosovitskiy, Fischer, Ilg, Hausser, Hazirbas,
  Golkov, van~der Smagt, Cremers, and Brox]{Dosovitskiy2015}
Alexey Dosovitskiy, Philipp Fischer, Eddy Ilg, Philip Hausser, Caner Hazirbas,
  Vladimir Golkov, Patrick van~der Smagt, Daniel Cremers, and Thomas Brox.
\newblock Flow{N}et: Learning optical flow with convolutional networks.
\newblock In \emph{The IEEE International Conference on Computer Vision
  (ICCV)}, pp.\  2758--2766, December 2015.

\bibitem[Finn et~al.(2016)Finn, Goodfellow, and Levine]{Finn2016}
Chelsea Finn, Ian Goodfellow, and Sergey Levine.
\newblock Unsupervised learning for physical interaction through video
  prediction.
\newblock In Daniel~D. Lee, Masashi Sugiyama, Ulrike von Luxburg, Isabelle
  Guyon, and Roman Garnett (eds.), \emph{Advances in Neural Information
  Processing Systems 29}, pp.\  64--72. Curran Associates, Inc., 2016.

\bibitem[Fourier(1822)]{Fourier1822}
Jean Baptiste~Joseph Fourier.
\newblock \emph{Théorie analytique de la chaleur}.
\newblock Didot, Firmin, 1822.

\bibitem[Fraccaro et~al.(2017)Fraccaro, Kamronn, Paquet, and
  Winther]{Fraccaro2017}
Marco Fraccaro, Simon Kamronn, Ulrich Paquet, and Ole Winther.
\newblock A disentangled recognition and nonlinear dynamics model for
  unsupervised learning.
\newblock In Isabelle Guyon, Ulrike von Luxburg, Samy Bengio, Hanna Wallach,
  Rob Fergus, S.~V.~N. Vishwanathan, and Roman Garnett (eds.), \emph{Advances
  in Neural Information Processing Systems 30}, pp.\  3601--3610. Curran
  Associates, Inc., 2017.

\bibitem[Franceschi et~al.(2020)Franceschi, Delasalles, Chen, Lamprier, and
  Gallinari]{Franceschi2020}
Jean-Yves Franceschi, Edouard Delasalles, Mickaël Chen, Sylvain Lamprier, and
  Patrick Gallinari.
\newblock Stochastic latent residual video prediction.
\newblock In Hal~Daumé III and Aarti Singh (eds.), \emph{Proceedings of the
  37th International Conference on Machine Learning}, volume 119 of
  \emph{Proceedings of Machine Learning Research}, pp.\  3233--3246. PMLR, July
  2020.

\bibitem[Goodfellow et~al.(2014)Goodfellow, Pouget-Abadie, Mirza, Xu,
  Warde-Farley, Ozair, Courville, and Bengio]{Goodfellow2014}
Ian Goodfellow, Jean Pouget-Abadie, Mehdi Mirza, Bing Xu, David Warde-Farley,
  Sherjil Ozair, Aaron Courville, and Yoshua Bengio.
\newblock Generative adversarial nets.
\newblock In Zoubin Ghahramani, Max Welling, Corinna Cortes, Neil~D. Lawrence,
  and Kilian~Q. Weinberger (eds.), \emph{Advances in Neural Information
  Processing Systems 27}, pp.\  2672--2680. Curran Associates, Inc., 2014.

\bibitem[Greydanus et~al.(2019)Greydanus, Dzamba, and Yosinski]{Greydanus2019}
Samuel Greydanus, Misko Dzamba, and Jason Yosinski.
\newblock Hamiltonian neural networks.
\newblock In Hanna Wallach, Hugo Larochelle, Alina Beygelzimer, Florence
  d'Alché Buc, Emily Fox, and Roman Garnett (eds.), \emph{Advances in Neural
  Information Processing Systems 32}, pp.\  15379--15389. Curran Associates,
  Inc., 2019.

\bibitem[Haber \& Ruthotto(2017)Haber and Ruthotto]{Haber2017}
Eldad Haber and Lars Ruthotto.
\newblock Stable architectures for deep neural networks.
\newblock \emph{Inverse Problems}, 34\penalty0 (1):\penalty0 014004, December
  2017.

\bibitem[Hairer et~al.(1993)Hairer, Nørsett, and Wanner]{Hairer1993}
Ernst Hairer, Syvert~P. Nørsett, and Gerhard Wanner.
\newblock \emph{Solving Ordinary Differential Equations I: Nonstiff Problems},
  chapter Runge-Kutta and Extrapolation Methods, pp.\  129--353.
\newblock Springer Berlin Heidelberg, Berlin, Heidelberg, 1993.

\bibitem[Hamilton(1835)]{Hamilton1835}
William~Rowan Hamilton.
\newblock Second essay on a general method in dynamics.
\newblock \emph{Philosophical Transactions of the Royal Society}, 125:\penalty0
  95--144, 1835.

\bibitem[He et~al.(2016)He, Zhang, Ren, and Sun]{He2016}
Kaiming He, Xiangyu Zhang, Shaoqing Ren, and Jian Sun.
\newblock Deep residual learning for image recognition.
\newblock In \emph{The IEEE Conference on Computer Vision and Pattern
  Recognition (CVPR)}, pp.\  770--778, June 2016.

\bibitem[Horn \& Schunck(1981)Horn and Schunck]{Horn1981}
Berthold K.~P. Horn and Brian~G. Schunck.
\newblock Determining optical flow.
\newblock \emph{Artificial Intelligence}, 17\penalty0 (1--3):\penalty0
  185--203, August 1981.

\bibitem[Hsieh et~al.(2018)Hsieh, Liu, Huang, Fei-Fei, and Niebles]{Hsieh2018}
Jun-Ting Hsieh, Bingbin Liu, De-An Huang, Li~Fei-Fei, and Juan~Carlos Niebles.
\newblock Learning to decompose and disentangle representations for video
  prediction.
\newblock In Samy Bengio, Hanna Wallach, Hugo Larochelle, Kristen Grauman,
  Nicolò Cesa-Bianchi, and Roman Garnett (eds.), \emph{Advances in Neural
  Information Processing Systems 31}, pp.\  517--526. Curran Associates, Inc.,
  2018.

\bibitem[Hsu et~al.(2017)Hsu, Zhang, and Glass]{Hsu2017}
Wei-Ning Hsu, Yu~Zhang, and James Glass.
\newblock Unsupervised learning of disentangled and interpretable
  representations from sequential data.
\newblock In Isabelle Guyon, Ulrike von Luxburg, Samy Bengio, Hanna Wallach,
  Rob Fergus, S.~V.~N. Vishwanathan, and Roman Garnett (eds.), \emph{Advances
  in Neural Information Processing Systems 30}, pp.\  1878--1889. Curran
  Associates, Inc., 2017.

\bibitem[Jaques et~al.(2020)Jaques, Burke, and Hospedales]{Jaques2020}
Miguel Jaques, Michael Burke, and Timothy Hospedales.
\newblock Physics-as-inverse-graphics: Unsupervised physical parameter
  estimation from video.
\newblock In \emph{International Conference on Learning Representations}, 2020.

\bibitem[Jia et~al.(2008)Jia, Xu, Zhao, and Li]{Jia2008}
Huabing Jia, Wei Xu, Xiaoshan Zhao, and Zhanguo Li.
\newblock Separation of variables and exact solutions to nonlinear diffusion
  equations with $x$-dependent convection and absorption.
\newblock \emph{Journal of Mathematical Analysis and Applications},
  339\penalty0 (2):\penalty0 982--995, March 2008.

\bibitem[Kalnins et~al.(1992)Kalnins, Miller, and Williams]{Kalnins1992}
E.~G. Kalnins, Willard Miller, Jr., and G.~C. Williams.
\newblock Recent advances in the use of separation of variables methods in
  general relativity.
\newblock \emph{Philosophical Transactions: Physical Sciences and Engineering},
  340\penalty0 (1658):\penalty0 337--352, 1992.

\bibitem[Kingma \& Ba(2015)Kingma and Ba]{Kingma2015}
Diederik~P. Kingma and Jimmy Ba.
\newblock Adam: A method for stochastic optimization.
\newblock In \emph{International Conference on Learning Representations}, 2015.

\bibitem[Kingma \& Welling(2014)Kingma and Welling]{Kingma2014}
Diederik~P. Kingma and Max Welling.
\newblock Auto-encoding variational {B}ayes.
\newblock In \emph{International Conference on Learning Representations}, 2014.

\bibitem[Kosiorek et~al.(2018)Kosiorek, Kim, Teh, and Posner]{Kosiorek2018}
Adam~R. Kosiorek, Hyunjik Kim, Yee~Whye Teh, and Ingmar Posner.
\newblock Sequential attend, infer, repeat: Generative modelling of moving
  objects.
\newblock In Samy Bengio, Hanna Wallach, Hugo Larochelle, Kristen Grauman,
  Nicolò Cesa-Bianchi, and Roman Garnett (eds.), \emph{Advances in Neural
  Information Processing Systems 31}, pp.\  8606--8616. Curran Associates,
  Inc., 2018.

\bibitem[Kraskov et~al.(2004)Kraskov, Stögbauer, and Grassberger]{Kraskov2004}
Alexander Kraskov, Harald Stögbauer, and Peter Grassberger.
\newblock Estimating mutual information.
\newblock \emph{Physical Review E}, 69:\penalty0 066138, June 2004.

\bibitem[Kutta(1901)]{Kutta1901}
Martin~Wilhelm Kutta.
\newblock Beitrag zur näherungweisen {I}ntegration totaler
  {D}ifferentialgleichungen.
\newblock \emph{Zeitschrift für Mathematik und Physik}, 45:\penalty0 435--453,
  1901.

\bibitem[Le~Dret \& Lucquin(2016)Le~Dret and Lucquin]{LeDret2016}
Hervé Le~Dret and Brigitte Lucquin.
\newblock \emph{Partial Differential Equations: Modeling, Analysis and
  Numerical Approximation}, chapter The Heat Equation, pp.\  219--251.
\newblock Springer International Publishing, Cham, 2016.

\bibitem[Le~Guen \& Thome(2020)Le~Guen and Thome]{LeGuen2020}
Vincent Le~Guen and Nicolas Thome.
\newblock Disentangling physical dynamics from unknown factors for unsupervised
  video prediction.
\newblock In \emph{The IEEE/CVF Conference on Computer Vision and Pattern
  Recognition (CVPR)}, pp.\  11474--11484, June 2020.

\bibitem[LeCun et~al.(1998)LeCun, Bottou, Bengio, and Haffner]{LeCun1998}
Yann LeCun, Léon Bottou, Yoshua Bengio, and Patrick Haffner.
\newblock Gradient-based learning applied to document recognition.
\newblock \emph{Proceedings of the IEEE}, 86\penalty0 (11):\penalty0
  2278--2324, November 1998.

\bibitem[Lee et~al.(2018)Lee, Zhang, Ebert, Abbeel, Finn, and Levine]{Lee2018}
Alex~X. Lee, Richard Zhang, Frederik Ebert, Pieter Abbeel, Chelsea Finn, and
  Sergey Levine.
\newblock Stochastic adversarial video prediction.
\newblock \emph{arXiv preprint arXiv:1804.01523}, 2018.

\bibitem[Li et~al.(2020)Li, Wong, Chen, and Duvenaud]{Li2020}
Xuechen Li, Ting-Kam~Leonard Wong, Ricky T.~Q. Chen, and David Duvenaud.
\newblock Scalable gradients for stochastic differential equations.
\newblock \emph{arXiv preprint arXiv:2001.01328}, 2020.

\bibitem[Liu et~al.(2019)Liu, Wu, Xu, Sun, Murphy, Freeman, and
  Tenenbaum]{Xu2019}
Zhijian Liu, Jiajun Wu, Zhenjia Xu, Chen Sun, Kevin Murphy, William~T. Freeman,
  and Joshua~B. Tenenbaum.
\newblock Modeling parts, structure, and system dynamics via predictive
  learning.
\newblock In \emph{International Conference on Learning Representations}, 2019.

\bibitem[Locatello et~al.(2019)Locatello, Bauer, Lucic, Rätsch, Gelly,
  Schölkopf, and Bachem]{Locatello2019}
Francesco Locatello, Stefan Bauer, Mario Lucic, Gunnar Rätsch, Sylvain Gelly,
  Bernhard Schölkopf, and Olivier Bachem.
\newblock Challenging common assumptions in the unsupervised learning of
  disentangled representations.
\newblock In Kamalika Chaudhuri and Ruslan Salakhutdinov (eds.),
  \emph{Proceedings of the 36th International Conference on Machine Learning},
  volume~97 of \emph{Proceedings of Machine Learning Research}, pp.\
  4114--4124, Long Beach, California, USA, June 2019. PMLR.

\bibitem[Long et~al.(2018)Long, Lu, Ma, and Dong]{Long2018}
Zichao Long, Yiping Lu, Xianzhong Ma, and Bin Dong.
\newblock {PDE}-{N}et: Learning {PDE}s from data.
\newblock In Jennifer Dy and Andreas Krause (eds.), \emph{Proceedings of the
  35th International Conference on Machine Learning}, volume~80 of
  \emph{Proceedings of Machine Learning Research}, pp.\  3208--3216,
  Stockholmsmässan, Stockholm Sweden, July 2018. PMLR.

\bibitem[Long et~al.(2019)Long, Lu, and Dong]{Long2019}
Zichao Long, Yiping Lu, and Bi~Dong.
\newblock {PDE}-{N}et 2.0: Learning {PDE}s from data with a numeric-symbolic
  hybrid deep network.
\newblock \emph{Journal of Computational Physics}, 399:\penalty0 108925, 2019.

\bibitem[Lu et~al.(2018)Lu, Zhong, Li, and Dong]{Lu2018}
Yiping Lu, Aoxiao Zhong, Quanzheng Li, and Bin Dong.
\newblock Beyond finite layer neural networks: Bridging deep architectures and
  numerical differential equations.
\newblock In Jennifer Dy and Andreas Krause (eds.), \emph{Proceedings of the
  35th International Conference on Machine Learning}, volume~80 of
  \emph{Proceedings of Machine Learning Research}, pp.\  3276--3285,
  Stockholmsmässan, Stockholm Sweden, July 2018. PMLR.

\bibitem[Madec \& Team()Madec and Team]{Madec2008}
Gurvan Madec and NEMO~System Team.
\newblock {NEMO} ocean engine.
\newblock Technical Report~27, Scientific Notes of Climate Modelling Center,
  Institut Pierre-Simon Laplace (IPSL). Zenodo.

\bibitem[Mathieu et~al.(2016)Mathieu, Couprie, and LeCun]{Mathieu2016}
Michael Mathieu, Camille Couprie, and Yann LeCun.
\newblock Deep multi-scale video prediction beyond mean square error.
\newblock In \emph{International Conference on Learning Representations}, 2016.

\bibitem[Micikevicius et~al.(2018)Micikevicius, Narang, Alben, Diamos, Elsen,
  Garcia, Ginsburg, Houston, Kuchaiev, Venkatesh, and Wu]{Micikevicius2018}
Paulius Micikevicius, Sharan Narang, Jonah Alben, Gregory Diamos, Erich Elsen,
  David Garcia, Boris Ginsburg, Michael Houston, Oleksii Kuchaiev, Ganesh
  Venkatesh, and Hao Wu.
\newblock Mixed precision training.
\newblock In \emph{International Conference on Learning Representations}, 2018.

\bibitem[Miller(1983)]{Miller1983}
Willard Miller, Jr.
\newblock The technique of variable separation for partial differential
  equations.
\newblock In Kurt~Bernardo Wolf (ed.), \emph{Nonlinear Phenomena}, pp.\
  184--208, Berlin, Heidelberg, 1983. Springer Berlin Heidelberg.

\bibitem[Miller(1988)]{Miller1988}
Willard Miller, Jr.
\newblock Mechanisms for variable separation in partial differential equations
  and their relationship to group theory.
\newblock In Decio Levi and Pavel Winternitz (eds.), \emph{Symmetries and
  Nonlinear Phenomena: Proceedings of the International School on Applied
  Mathematics}, pp.\  188--221, Singapore, 1988. World Scientific.

\bibitem[Minderer et~al.(2019)Minderer, Sun, Villegas, Cole, Murphy, and
  Lee]{Minderer2019}
Matthias Minderer, Chen Sun, Ruben Villegas, Forrester Cole, Kevin Murphy, and
  Honglak Lee.
\newblock Unsupervised learning of object structure and dynamics from videos.
\newblock In Hanna Wallach, Hugo Larochelle, Alina Beygelzimer, Florence
  d'Alché Buc, Emily Fox, and Roman Garnett (eds.), \emph{Advances in Neural
  Information Processing Systems 32}, pp.\  92--102. Curran Associates, Inc.,
  2019.

\bibitem[Paszke et~al.(2019)Paszke, Gross, Massa, Lerer, Bradbury, Chanan,
  Killeen, Lin, Gimelshein, Antiga, Desmaison, Kopf, Yang, DeVito, Raison,
  Tejani, Chilamkurthy, Steiner, Fang, Bai, and Chintala]{Paszke2019}
Adam Paszke, Sam Gross, Francisco Massa, Adam Lerer, James Bradbury, Gregory
  Chanan, Trevor Killeen, Zeming Lin, Natalia Gimelshein, Luca Antiga, Alban
  Desmaison, Andreas Kopf, Edward Yang, Zachary DeVito, Martin Raison, Alykhan
  Tejani, Sasank Chilamkurthy, Benoit Steiner, Lu~Fang, Junjie Bai, and Soumith
  Chintala.
\newblock Py{T}orch: An imperative style, high-performance deep learning
  library.
\newblock In Hanna Wallach, Hugo Larochelle, Alina Beygelzimer, Florence
  d'Alché Buc, Emily Fox, and Roman Garnett (eds.), \emph{Advances in Neural
  Information Processing Systems 32}, pp.\  8026--8037. Curran Associates,
  Inc., 2019.

\bibitem[Polyanin(2019)]{Polyanin2019}
Andrei~D. Polyanin.
\newblock Functional separable solutions of nonlinear convection–diffusion
  equations with variable coefficients.
\newblock \emph{Communications in Nonlinear Science and Numerical Simulation},
  73:\penalty0 379--390, July 2019.

\bibitem[Polyanin(2020)]{Polyanin2020a}
Andrei~D. Polyanin.
\newblock Functional separation of variables in nonlinear {PDEs}: General
  approach, new solutions of diffusion-type equations.
\newblock \emph{Mathematics}, 8\penalty0 (1):\penalty0 90, 2020.

\bibitem[Polyanin \& Zhurov(2020)Polyanin and Zhurov]{Polyanin2020b}
Andrei~D. Polyanin and Alexei~I. Zhurov.
\newblock Separation of variables in {PDEs} using nonlinear transformations:
  Applications to reaction–diffusion type equations.
\newblock \emph{Applied Mathematics Letters}, 100:\penalty0 106055, February
  2020.

\bibitem[Radford et~al.(2016)Radford, Metz, and Chintala]{Radford2016}
Alec Radford, Luke Metz, and Soumith Chintala.
\newblock Unsupervised representation learning with deep convolutional
  generative adversarial networks.
\newblock In \emph{International Conference on Learning Representations}, 2016.

\bibitem[Raissi(2018)]{Raissi2018}
Maziar Raissi.
\newblock Deep hidden physics models: Deep learning of nonlinear partial
  differential equations.
\newblock \emph{Journal of Machine Learning Research}, 19\penalty0
  (25):\penalty0 1--24, 2018.

\bibitem[Raissi et~al.(2020)Raissi, Yazdani, and Karniadakis]{Raissi2020}
Maziar Raissi, Alireza Yazdani, and George~Em Karniadakis.
\newblock Hidden fluid mechanics: Learning velocity and pressure fields from
  flow visualizations.
\newblock \emph{Science}, 367\penalty0 (6481):\penalty0 1026--1030, 2020.

\bibitem[Rezende et~al.(2014)Rezende, Mohamed, and Wierstra]{Rezende2014}
Danilo~Jimenez Rezende, Shakir Mohamed, and Daan Wierstra.
\newblock Stochastic backpropagation and approximate inference in deep
  generative models.
\newblock In Eric~P. Xing and Tony Jebara (eds.), \emph{Proceedings of the 31st
  International Conference on Machine Learning}, volume~32 of \emph{Proceedings
  of Machine Learning Research}, pp.\  1278--1286, Bejing, China, June 2014.
  PMLR.

\bibitem[Ronneberger et~al.(2015)Ronneberger, Fischer, and
  Brox]{Ronneberger2015}
Olaf Ronneberger, Philipp Fischer, and Thomas Brox.
\newblock U-net: Convolutional networks for biomedical image segmentation.
\newblock In Nassir Navab, Joachim Hornegger, William~M. Wells, and
  Alejandro~F. Frangi (eds.), \emph{Medical Image Computing and
  Computer-Assisted Intervention -- MICCAI 2015}, pp.\  234--241, Cham, 2015.
  Springer International Publishing.

\bibitem[Rubanova et~al.(2019)Rubanova, Chen, and Duvenaud]{Rubanova2019}
Yulia Rubanova, Ricky T.~Q. Chen, and David Duvenaud.
\newblock Latent ordinary differential equations for irregularly-sampled time
  series.
\newblock In Hanna Wallach, Hugo Larochelle, Alina Beygelzimer, Florence
  d'Alché Buc, Emily Fox, and Roman Garnett (eds.), \emph{Advances in Neural
  Information Processing Systems 32}, pp.\  5320--5330. Curran Associates,
  Inc., 2019.

\bibitem[Ryder et~al.(2018)Ryder, Golightly, McGough, and Prangle]{Ryder2018}
Tom Ryder, Andrew Golightly, A.~Stephen McGough, and Dennis Prangle.
\newblock Black-box variational inference for stochastic differential
  equations.
\newblock In Jennifer Dy and Andreas Krause (eds.), \emph{Proceedings of the
  35th International Conference on Machine Learning}, volume~80 of
  \emph{Proceedings of Machine Learning Research}, pp.\  4423--4432,
  Stockholmsmässan, Stockholm Sweden, July 2018. PMLR.

\bibitem[Saha et~al.(2020)Saha, Dash, and Mukhopadhyay]{Saha2020}
Priyabrata Saha, Saurabh Dash, and Saibal Mukhopadhyay.
\newblock Ph{IC}net: Physics-incorporated convolutional recurrent neural
  networks for modeling dynamical systems.
\newblock \emph{arXiv preprint arXiv:2004.06243}, 2020.

\bibitem[Schüldt et~al.(2004)Schüldt, Laptev, and Caputo]{Schuldt2004}
Christian Schüldt, Ivan Laptev, and Barbara Caputo.
\newblock Recognizing human actions: A local {SVM} approach.
\newblock In \emph{Proceedings of the 17th International Conference on Pattern
  Recognition, 2004. ICPR 2004.}, volume~3, pp.\  32--36, August 2004.

\bibitem[Shi et~al.(2015)Shi, Chen, Wang, Yeung, Wong, and Woo]{Shi2015}
Xingjian Shi, Zhourong Chen, Hao Wang, Dit-Yan Yeung, Wai-kin Wong, and
  Wang-chun Woo.
\newblock Convolutional {LSTM} network: A machine learning approach for
  precipitation nowcasting.
\newblock In Corinna Cortes, Neil~D. Lawrence, Daniel~D. Lee, Masashi Sugiyama,
  and Roman Garnett (eds.), \emph{Advances in Neural Information Processing
  Systems 28}, pp.\  802--810. Curran Associates, Inc., 2015.

\bibitem[Simonyan \& Zisserman(2015)Simonyan and Zisserman]{Simonyan2015}
Karen Simonyan and Andrew Zisserman.
\newblock Very deep convolutional networks for large-scale image recognition.
\newblock In \emph{International Conference on Learning Representations}, 2015.

\bibitem[Sirignano \& Spiliopoulos(2018)Sirignano and
  Spiliopoulos]{Sirignano2018}
Justin Sirignano and Konstantinos Spiliopoulos.
\newblock {DGM}: A deep learning algorithm for solving partial differential
  equations.
\newblock \emph{Journal of Computational Physics}, 375:\penalty0 1339--1364,
  2018.

\bibitem[Srivastava et~al.(2015)Srivastava, Mansimov, and
  Salakhudinov]{Srivastava2015}
Nitish Srivastava, Elman Mansimov, and Ruslan Salakhudinov.
\newblock Unsupervised learning of video representations using {LSTM}s.
\newblock In Francis Bach and David Blei (eds.), \emph{Proceedings of the 32nd
  International Conference on Machine Learning}, volume~37 of \emph{Proceedings
  of Machine Learning Research}, pp.\  843--852, Lille, France, July 2015.
  PMLR.

\bibitem[Tompson et~al.(2017)Tompson, Schlachter, Sprechmann, and
  Perlin]{Tompson2017}
Jonathan Tompson, Kristofer Schlachter, Pablo Sprechmann, and Ken Perlin.
\newblock Accelerating {E}ulerian fluid simulation with convolutional networks.
\newblock In Doina Precup and Yee~Whye Teh (eds.), \emph{Proceedings of the
  34th International Conference on Machine Learning}, volume~70 of
  \emph{Proceedings of Machine Learning Research}, pp.\  3424--3433,
  International Convention Centre, Sydney, Australia, August 2017. PMLR.

\bibitem[Toth et~al.(2020)Toth, Rezende, Jaegle, Racanière, Botev, and
  Higgins]{Toth2020}
Peter Toth, Danilo~J. Rezende, Andrew Jaegle, Sébastien Racanière, Aleksandar
  Botev, and Irina Higgins.
\newblock Hamiltonian generative networks.
\newblock In \emph{International Conference on Learning Representations}, 2020.

\bibitem[Tulyakov et~al.(2018)Tulyakov, Liu, Yang, and Kautz]{Tulyakov2018}
Sergey Tulyakov, Ming-Yu Liu, Xiaodong Yang, and Jan Kautz.
\newblock Mo{C}o{GAN}: Decomposing motion and content for video generation.
\newblock In \emph{The IEEE Conference on Computer Vision and Pattern
  Recognition (CVPR)}, pp.\  1526--1535, June 2018.

\bibitem[Unterthiner et~al.(2018)Unterthiner, van Steenkiste, Kurach, Marinier,
  Michalski, and Gelly]{Unterthiner2018}
Thomas Unterthiner, Sjoerd van Steenkiste, Karol Kurach, Raphaël Marinier,
  Marcin Michalski, and Sylvain Gelly.
\newblock Towards accurate generative models of video: A new metric \&
  challenges.
\newblock \emph{arXiv preprint arXiv:1812.01717}, 2018.

\bibitem[van Steenkiste et~al.(2018)van Steenkiste, Chang, Greff, and
  Schmidhuber]{Steenkiste2018}
Sjoerd van Steenkiste, Michael Chang, Klaus Greff, and Jürgen Schmidhuber.
\newblock Relational neural expectation maximization: Unsupervised discovery of
  objects and their interactions.
\newblock In \emph{International Conference on Learning Representations}, 2018.

\bibitem[Villegas et~al.(2017{\natexlab{a}})Villegas, Yang, Hong, Lin, and
  Lee]{Villegas2017a}
Ruben Villegas, Jimei Yang, Seunghoon Hong, Xunyu Lin, and Honglak Lee.
\newblock Decomposing motion and content for natural video sequence prediction.
\newblock In \emph{International Conference on Learning Representations},
  2017{\natexlab{a}}.

\bibitem[Villegas et~al.(2017{\natexlab{b}})Villegas, Yang, Zou, Sohn, Lin, and
  Lee]{Villegas2017b}
Ruben Villegas, Jimei Yang, Yuliang Zou, Sungryull Sohn, Xunyu Lin, and Honglak
  Lee.
\newblock Learning to generate long-term future via hierarchical prediction.
\newblock In Doina Precup and Yee~Whye Teh (eds.), \emph{Proceedings of the
  34th International Conference on Machine Learning}, volume~70 of
  \emph{Proceedings of Machine Learning Research}, pp.\  3560--3569,
  International Convention Centre, Sydney, Australia, August
  2017{\natexlab{b}}. PMLR.

\bibitem[Villegas et~al.(2019)Villegas, Pathak, Kannan, Erhan, Le, and
  Lee]{Villegas2019}
Ruben Villegas, Arkanath Pathak, Harini Kannan, Dumitru Erhan, Quoc~V. Le, and
  Honglak Lee.
\newblock High fidelity video prediction with large stochastic recurrent neural
  networks.
\newblock In Hanna Wallach, Hugo Larochelle, Alina Beygelzimer, Florence
  d'Alché Buc, Emily Fox, and Roman Garnett (eds.), \emph{Advances in Neural
  Information Processing Systems 32}, pp.\  81--91. Curran Associates, Inc.,
  2019.

\bibitem[Vondrick et~al.(2016)Vondrick, Pirsiavash, and Torralba]{Vondrick2016}
Carl Vondrick, Hamed Pirsiavash, and Antonio Torralba.
\newblock Generating videos with scene dynamics.
\newblock In Daniel~D. Lee, Masashi Sugiyama, Ulrike von Luxburg, Isabelle
  Guyon, and Roman Garnett (eds.), \emph{Advances in Neural Information
  Processing Systems 29}, pp.\  613--621. Curran Associates, Inc., 2016.

\bibitem[Wang et~al.(2017)Wang, Long, Wang, Gao, and Yu]{Wang2017}
Yunbo Wang, Mingsheng Long, Jianmin Wang, Zhifeng Gao, and Philip~S Yu.
\newblock Pred{RNN}: Recurrent neural networks for predictive learning using
  spatiotemporal {LSTM}s.
\newblock In Isabelle Guyon, Ulrike von Luxburg, Samy Bengio, Hanna Wallach,
  Rob Fergus, S.~V.~N. Vishwanathan, and Roman Garnett (eds.), \emph{Advances
  in Neural Information Processing Systems 30}, pp.\  879--888. Curran
  Associates, Inc., 2017.

\bibitem[Wang et~al.(2018)Wang, Gao, Long, Wang, and Yu]{Wang2018}
Yunbo Wang, Zhifeng Gao, Mingsheng Long, Jianmin Wang, and Philip~S Yu.
\newblock {P}red{RNN}++: Towards a resolution of the deep-in-time dilemma in
  spatiotemporal predictive learning.
\newblock In Jennifer Dy and Andreas Krause (eds.), \emph{Proceedings of the
  35th International Conference on Machine Learning}, volume~80 of
  \emph{Proceedings of Machine Learning Research}, pp.\  5123--5132,
  Stockholmsmässan, Stockholm Sweden, July 2018. PMLR.

\bibitem[Wang et~al.(2019{\natexlab{a}})Wang, Jiang, Yang, Li, Long, and
  Fei-Fei]{Wang2019b}
Yunbo Wang, Lu~Jiang, Ming-Hsuan Yang, Li-Jia Li, Mingsheng Long, and
  Li~Fei-Fei.
\newblock Eidetic {3D} {LSTM}: A model for video prediction and beyond.
\newblock In \emph{International Conference on Learning Representations},
  2019{\natexlab{a}}.

\bibitem[Wang et~al.(2019{\natexlab{b}})Wang, Zhang, Zhu, Long, Wang, and
  Yu]{Wang2019}
Yunbo Wang, Jianjin Zhang, Hongyu Zhu, Mingsheng Long, Jianmin Wang, and
  Philip~S. Yu.
\newblock Memory in memory: A predictive neural network for learning
  higher-order non-stationarity from spatiotemporal dynamics.
\newblock In \emph{The IEEE/CVF Conference on Computer Vision and Pattern
  Recognition (CVPR)}, June 2019{\natexlab{b}}.

\bibitem[Weissenborn et~al.(2020)Weissenborn, Täckström, and
  Uszkoreit]{Weissenborn2020}
Dirk Weissenborn, Oscar Täckström, and Jakob Uszkoreit.
\newblock Scaling autoregressive video models.
\newblock In \emph{International Conference on Learning Representations}, 2020.

\bibitem[Yingzhen \& Mandt(2018)Yingzhen and Mandt]{Yingzhen2018}
Li~Yingzhen and Stephan Mandt.
\newblock Disentangled sequential autoencoder.
\newblock In Jennifer Dy and Andreas Krause (eds.), \emph{Proceedings of the
  35th International Conference on Machine Learning}, volume~80 of
  \emph{Proceedings of Machine Learning Research}, pp.\  5670--5679,
  Stockholmsmässan, Stockholm Sweden, July 2018. PMLR.

\bibitem[Yıldız et~al.(2019)Yıldız, Heinonen, and Lahdesmaki]{Yildiz2019}
Cagatay Yıldız, Markus Heinonen, and Harri Lahdesmaki.
\newblock {ODE}$^2${VAE}: Deep generative second order odes with {B}ayesian
  neural networks.
\newblock In Hanna Wallach, Hugo Larochelle, Alina Beygelzimer, Florence
  d'Alché Buc, Emily Fox, and Roman Garnett (eds.), \emph{Advances in Neural
  Information Processing Systems 32}, pp.\  13412--13421. Curran Associates,
  Inc., 2019.

\bibitem[Zhang et~al.(2017)Zhang, Zheng, and Qi]{Zhang2017}
Junbo Zhang, Yu~Zheng, and Dekang Qi.
\newblock Deep spatio-temporal residual networks for citywide crowd flows
  prediction.
\newblock In \emph{Proceedings of the Thirty-First AAAI Conference on
  Artificial Intelligence}, AAAI'17, pp.\  1655--1661. AAAI Press, 2017.

\end{thebibliography}
\bibliographystyle{template/iclr2021_conference}

\appendix

\section{Proofs}

\subsection{Resolution of the Heat Equation}
\label{sec:heateqdetails}

In this section, we succinctly detail a proof for the existence and uniqueness for the solution to the two-dimensional heat equation.
It shows that product-separable solutions allow to build the entire solution space for this problem, highlighting our interest in the research of separable solutions.

\paragraph{Existence through separation of variables.}
Consider the heat equation problem: 
\begin{align}
    \label{eq.heat_supp}
    \dpd{u}{t} = c^{2} \dpd[2]{u}{x}, && u\parentheses*{0, t} = u\parentheses*{L, t} = 0, && u\parentheses*{x, 0} = f\parentheses*{x}.
\end{align}

Assuming product separability of $u$ with $u\parentheses*{x,t}=u_1\parentheses*{x} u_2\parentheses*{t}$ in \cref{eq.heat_supp} gives: 
\begin{equation}
    \label{eq:constant}
    c^{2} \frac{u_{1}''\parentheses*{x}}{u_{1}\parentheses*{x}} = \frac{u_{2}'\parentheses*{t}}{u_{2}\parentheses*{t}}.
\end{equation}
Both sides being independent of each other variables, they are equal to a constant denoted by $- \alpha$.
If $\alpha$ is negative, solving the right side of \cref{eq:constant} results to non-physical solutions with exponentially increasing temperatures, and imposing border condition of \cref{eq.heat_supp} makes this solution collapse to the null trivial solution.
Therefore, we consider that $\alpha > 0$. 

Both sides of \cref{eq:constant} being equal to a constant leads to a second-order ODE on $u_1$ and a first-order ODE on $u_2$, giving the solution shapes, with constants $A$, $B$ and $D$:
\begin{equation}
    \begin{cases}
        u_1\parentheses*{x} & = A \cos\parentheses*{\sqrt{\alpha} x} + B \sin\parentheses*{\sqrt{\alpha} x}\\
        u_2\parentheses*{t} & = D \erm^{-\alpha c^{2} t}
    \end{cases}.
\end{equation}

\paragraph{Link with initial and boundary conditions.}
We now link the above equation to the boundary conditions of the problem.
Because our separation is multiplicative, we can omit $D$ for non-trivial solutions and set it without loss of generality to $1$, as it only scales the values of $A$ and $B$.

Boundary condition $u\parentheses*{0,t} = u\parentheses*{L,t}=0$, along with the fact that for all $t > 0$, $u_{2}\parentheses*{t} \neq 0$, give: 
\begin{align}
    &A = 0, && &B \erm^{-\alpha c^{2} t} \sin\parentheses*{\sqrt{\alpha}L}=0,
\end{align} 
which means that, for a non-trivial solution (i.e., $B \neq 0$), we have for some $n \in \mathbb{N}$: $\sqrt{\alpha} = n \pi / L$.
We can finally express our product-separable solution to the heat equation without initial conditions as:
\begin{equation}
    \label{heat-sol}
    u\parentheses*{x, t} = B \sin\parentheses*{\frac{n \pi}{L} x} \exp\parentheses*{- \parentheses*{\frac{c n \pi}{L}}^{2} t}. 
\end{equation}

Considering the superposition principle, because the initial problem is homogeneous, all linear combinations of \cref{heat-sol} are solutions of the heat equation without initial conditions.
Therefore, any following function is a solution of the heat equation without initial conditions.
\begin{equation}
    u\parentheses*{x, t} = \sum_{n=0}^{+\infty} B_{n} \sin\parentheses*{\frac{n \pi}{L} x} \exp\parentheses*{- \parentheses*{\frac{c n \pi}{L}}^{2} t}. \label{fourier}
\end{equation}

Finally, considering the initial condition $u\parentheses*{x, 0} = f\parentheses*{x}$, a Fourier decomposition of $f$ allows to choose appropriate values for all coefficients $B_n$, showing that, for any initial condition $f$, there exists a solution to \cref{eq.heat_supp} of the form of \cref{fourier}.

\paragraph{Uniqueness.}
We present here elements of proof for establishing the uniqueness of the solutions of \cref{eq.heat_supp} that belong to $\gC^2\parentheses*{\brackets*{0, 1} \times \sR_{+}}$.
Detailed and rigorous proofs are given by \citet{LeDret2016}.

The key element consists in establishing the so-called Maximum Principle which states that, considering a sufficiently smooth solution, the minimum value of the solution is reached on the boundary of the space and time domains.

For null border condition as in our case, this means that the norm of the solution $u$ is given by the norm of the initial condition $f$.
Finally, let us consider two smooth solutions $U_{1}$ and $U_{2}$ of \cref{eq.heat_supp}.
Then, their difference $v = U_{1} - U_{2}$ follows the heat equation with null border and initial conditions (i.e, $v\parentheses*{x, 0}=0$).
Because $v$ is as regular as $U_{1}$ and $U_{2}$, it satisfies the previous fact about the norm of the solutions, i.e, the norm of $v$ equals the norm of its initial condition:  $\euclideannorm*{v}=0$.
Therefore, $v$ is null and so is $U_{1} - U_{2} = 0$, showing the uniqueness of the solutions.

Therefore, this shows that solutions of the form of \cref{fourier} shape the whole set of smooth solutions of \cref{eq.heat_supp}.

\subsection{Heat Equation with Advection Term}
\label{sec:heateq2details}

Consider the heat equation with a complementary advection term, for $x \in \parentheses*{-1, 1}$, $t \in \parentheses*{0, T}$ and a constant $c \in \sR_{+}$.
\begin{equation}
    \pd{u}{t} + c \pd{u}{x} = \chi \pd[2]{u}{x}, \hspace{0.1in}.
     \label{eq:heateq2}
\end{equation}

We give here details for the existence of product-separable solutions of \cref{eq:heateq2}.
To this end, let us choose real constants $\alpha$ and $\beta$, and consider the following change of variables for $u$:
\begin{equation}
    u\parentheses*{x,t} = v\parentheses*{x,t} \erm^{\alpha x + \beta t}.
\end{equation}

The partial derivatives from \cref{eq:heateq2} can be rewritten as functions of the new variable $v$:
\begin{align}
    \dpd{u}{t} & = \dpd{v}{t} \erm^{\alpha x + \beta t} + \beta v \erm^{\alpha x + \beta t} \\
    \dpd{u}{x} &= \dpd{v}{x} \erm^{\alpha x + \beta t} + \alpha v \erm^{\alpha x + \beta t} \\
    \dpd[2]{u}{x} & = \dpd[2]{v}{x} \erm^{\alpha x + \beta t} + 2 \alpha \dpd{v}{x} \erm^{\alpha x + \beta t} + \alpha^2 v \erm^{\alpha x + \beta t} 
\end{align}
Using these expressions in \cref{eq:heateq2} and dividing it by $\erm^{\alpha x + \beta t}$ lead to:
\begin{equation}
    \pd{v}{t} + \parentheses*{\beta + c\alpha-\alpha^2 \chi} v + \parentheses*{c-2\alpha \chi} \pd{v}{x} = \nu \pd[2]{v}{x}.
\end{equation}
$\alpha$ and $\beta$ can then be set such that:
\begin{align}
    \beta + c\alpha - \alpha^2 \chi = 0 && c - 2\alpha \chi = 0,
\end{align}
to retrieve the standard two-dimensional heat equation of \cref{eq.heat_supp} given by:
\begin{equation}
    \pd{v}{t} = \chi \pd[2]{v}{x},
\end{equation}
which is known to have product-separable solutions as explained in the previous section.
This more generally shows that all solutions of \cref{eq:heateq2} can be retrieved from solutions to \cref{eq.heat_supp}.

\section{Accessing Time Derivatives of $S$ and Deriving a Feasible Weaker Constraint}
\label{sec.derivative}

Explicitly constraining the time derivative of $E_{S}\parentheses*{V_{\tau}\parentheses*{t}}$ as explained in \cref{sec:disentanglement} is a difficult matter in practice.
Indeed, $E_{S}$ does not take as input neither the time coordinate $t$ nor spatial coordinates $x$ and $y$ as done by \cite{Raissi2018} and \cite{Sirignano2018}, which allows them to directly estimate the networks derivative thanks to automatic differentiation.
In our case, $E_{S}$ rather takes as input a finite number of observations, making this derivative impractical to compute.

To discretize \cref{s_invar_strict} and find a weaker constraint, we chose to leverage the Cauchy-Schwarz inequality.
We presented and used a version where we applied this inequality on the whole integration domain, i.e., from $t_{0}$ to $t_{1} - \tau \Delta t$.
We highlight that this inequality can also be applied on subintervals of the integration domain, generalizing our proposition.
Indeed, let $p \in \sN^{\ast}$ and consider a sequence of $t^{\parentheses*{k}}$ for $k \in \lrbrackets*{0, p}$ such that $t_{0} = t^{\parentheses*{0}} \leq t^{\parentheses*{1}} \leq \ldots \leq t^{\parentheses*{p}} = t_{1} - \tau \Delta t$.
Then, using the Cauchy-Schwarz inequality, we obtain:
\begin{equation}
    \begin{aligned}
        \int_{t_{0}}^{t_{1} - \tau \Delta t} \euclideannorm*{\dpd{E_{S}\parentheses*{V_{\tau}\parentheses*{t}}}{t}}_{2}^{2} \dif t & = \sum_{k = 0}^{k = p} \int_{t^{\parentheses*{k - 1}}}^{t^{\parentheses*{k}}} \euclideannorm*{\dpd{E_{S}\parentheses*{V_{\tau}\parentheses*{t}}}{t}}_{2}^{2} \dif t \\
        & \geq \sum_{k = 0}^{k=p} \frac{1}{t^{\parentheses*{k}} - t^{\parentheses*{k - 1}}} \euclideannorm*{\int_{t^{\parentheses*{k - 1}}}^{t^{\parentheses*{k}}} \dpd{E_{S}\parentheses*{V_{\tau}\parentheses*{t}}}{t} \dif t}_{2}^{2} \\
        & \geq \sum_{k = 0}^{k = p} \frac{1}{t^{\parentheses*{k}} - t^{\parentheses*{k - 1}}} \euclideannorm*{E_{S}\parentheses*{V_{\tau}\parentheses*{t^{\parentheses*{k}}}} - E_{S}\parentheses*{V_{\tau}\parentheses*{t^{\parentheses*{k - 1}}}}
        }_{2}^{2}.
    \end{aligned}
\end{equation}
Our constraint is a special case of this development, with $p = 1$.
Nevertheless, we experimentally found that our simple penalty is sufficient to achieve state-of-the-art performance at a substantially reduced computational cost.
We notice that other invariance constraints such as the one of \cite{Denton2017} can also be derived thanks to framework, showing the generality of our approach.

\section{Of Spatiotemporal Disentanglement}
\label{app.disentanglement}

\subsection{Modeling Spatiotemporal Phenomena with Differential Equations}
\label{app:DifferentialEquationsDiscussion}

Besides their increasing popularity to model spatiotemporal phenomena (see \cref{sec:related}), the ability of residual networks to facilitate learning \citep{Haber2017} as well as the success of their continuous counterpart \citep{Chen2018} motivate our choice.
Indeed, learning ODEs or discrete approximations as residual networks has become standard for a variety of tasks such as classification, inpainting, and generative models.
Consequently, their application to forecasting physical processes and videos is only a natural extension of its already broad applicability discussed in \cref{sec:related}.
Furthermore, they present interesting properties, as detailed below.

\subsection{Separation of Variables Preserves the Mutual Information of S and T through Time}
\label{sec.suppl_MI}

\subsubsection{Invertible Flow of an ODE}

We first highlight that the general ODE \cref{eq.time_ode} admits, according to the Cauchy–Lipschitz theorem, exactly one solution for a given initial condition, since $f$ is implemented with a standard neural network (see \cref{sec:training}), making it Lipschitz-continuous.
Consequently, the flow of this ODE, denoted by $\Phi_{t}$ and defined as:
\begin{align*}
    \Phi \colon \sR \times \sR^{p} & \to \sR^{p} \\
    \parentheses*{t_{0}, T_{t_{0}}} & \mapsto \Phi_{t}\parentheses*{T_{t_{0}}} = T_{t_{0} + t}
\end{align*}
is a bijection for all $t$.
Indeed, let $T_{t_{0}}$ be fixed and $t_{0}$, $t_{1}$ be two timesteps; thanks to the existence and unicity of the solution to the ODE with this initial condition: $\Phi_{t_{0} + t_{1}} = \Phi_{t_{0}} \circ \Phi_{t_{1}} = \Phi_{t_{1}} \circ \Phi_{t_{0}}$.
Therefore, $\Phi_{t}$ is a bijection and $\Phi_{t}^{-1} = \Phi_{-t}$.
Moreover, the flow is differentiable if $f$ is continuously differentiable as well, which is not a restrictive assumption if it is implemented by a neural network with differentiable activation functions.

\subsubsection{Preservation of Mutual Information by Invertible Mappings}

A proof of the following result is given by \cite{Kraskov2004}.
We indicate below the major steps of the proof.
Let $X$ and $Y$ be two random variables with marginal densities $\mu_{X}$, $\mu_{Y}$.
Let $F$ be a diffeomorphism acting on $Y$, $Y' = F\parentheses*{Y}$.
If $J_{F}$ is the determinant of the Jacobian of $F$, we have:
\begin{equation*}
    \mu'\parentheses*{x, y'} = \mu\parentheses*{x, y} J_{F}\parentheses*{y'}.
\end{equation*}
Then, expressing the mutual information $I$ in integral form, with the change of variables $y' = F\parentheses*{y}$ ($F$ being a diffeomorphism), results in:
\begin{align*}
    I\parentheses*{X, Y'} &= \iint_{x, y'} \mu'\parentheses*{x, y'} \log \frac{\mu'\parentheses*{x, y'}}{\mu_{X}\parentheses*{x} \times \mu_{Y'}\parentheses*{y'}} \dif x \dif y' \\
    & = \iint_{x, y} \mu\parentheses*{x, y}  \log \frac{\mu\parentheses*{x, y}}{\mu_{X}\parentheses*{x} \times \mu_{Y}\parentheses*{y}} \dif x \dif y \\
    I\parentheses*{X, Y'} & = I\parentheses*{X, Y}.
\end{align*}

\subsection{Ensuring Disentanglement at any Time}

As noted by \cite{Chen2016} and \cite{Achille2018}, mutual information $I$ is a key metric to evaluate disentanglement.
We show that our model logically preserves the mutual information between $S$ and $T$ through time thanks to the flow of the learned ODE on $T$.
Indeed, with the result of mutual information preservation by diffeomorphisms, and $\Phi_{t}$ being a diffeomorphism as demonstrated above, we have, for all $t$ and $t'$:
\begin{equation}
    I\parentheses*{S, T_{t}} = I\parentheses*{X, \Phi_{t' - t}\parentheses*{T_{t}}} = I\parentheses*{S, T_{t'}}.
\end{equation}
Hence, if $S$ and $T_{t}$ are disentangled, then so are $S$ and $T_{t'}$.

The flow $\Phi_{t}$ being dicretized in practice, its invertibility can no longer be guaranteed in general.
Some numerical schemes \citep{Chen2020} or residual networks with Lipschitz-constrained residual blocks \citep{Behrmann2019} provide sufficient conditions to concretely reach this invertibility.
In our case, we did not observe the need to enforce invertibility.
We can also leverage the data processing inequality to show that, for any $t \geq t_{0}$:
\begin{equation}
    I\parentheses*{S, T_{t_{0}}} \geq I\parentheses*{S, T_{t}},
\end{equation}
since $T_{t}$ is a deterministic function of $T_{t_{0}}$.
Since we constrain the very first $T$ value $T_{t_{0}}$ (i.e., we do not need to go back in time), there is no imperative need to enforce the invertibility of $\Phi_{t}$ in practice: the inequality also implies that, if $S$ and $T_{t_{0}}$ are disentangled, then so are $S$ and $T_{t}$ for $t \geq t_{0}$.
Nevertheless, should the need to disentangle for $t < t_{0}$ appear, the aforementioned mutual information conservation properties could allow, with further practical work to ensure the effective invertibility of $\Phi_{t}$, to still regularize $T_{t_{0}}$ only.
This is, however, out of the scope of this paper.

\section{Datasets}
\label{sec:datasets}

\subsection{WaveEq and WaveEq-100}

These datasets are based on the two-dimensional wave equation on a functional $w\parentheses*{x, y, t}$:
\begin{equation}
    \label{eq.wave_eq}
    \pd[2]{w}{t} = c^2 \nabla^{2} w + f\parentheses*{x, y, t},
\end{equation}
where $\nabla^{2}$ is the Laplacian operator, $c$ denotes the wave celerity, and $f$ is an arbitrary time-dependent source term.
It has several application in physics, modeling a wide range of phenomena ranging from mechanical oscillations to electromagnetism.
Note that the homogeneous equation, where $f = 0$, admits product-separable solutions.

We build the WaveEq dataset by solving \cref{eq.wave_eq} for $t \in \brackets*{0, 0.298}$ and $x, y \in \brackets*{0, 63}$.
Sequences are generated using $c$ drawn uniformly at random in $\brackets*{300, 400}$ for each sequence to imitate the propagation of acoustic waves, with initial and Neumann boundary conditions:
\begin{equation}
    w\parentheses*{x, y, 0} = w\parentheses*{0, 0, t} = w\parentheses*{32, 32, t} = 0,
\end{equation}
and, following \cite{Saha2020}, we make use of the following source term:
\begin{equation}
    f\parentheses{x, y, t} =
    \begin{cases}
        f_{0} \erm^{-\frac{t}{T_{0}}} & \mbox{if $\parentheses*{x, y} \in \gB\parentheses*{\parentheses*{32, 32}, 5}$} \\
        0 & \mbox{otherwise}
    \end{cases},
\end{equation}
with $T_{0} = 0.05$ and $f_{0} \sim \gU\parentheses*{\brackets*{1, 30}}$.
The source term is taken non-null in a circular central zone only in order to avoid numerical differentiation problems in the case of a punctual source.

We generate $300$ sequences of $64 \times 64$ frames of length $150$ from this setting by assimilating pixel $\parentheses*{i, j} \in \lrbrackets*{0, 63} \times \lrbrackets*{0, 63}$ to a point $\parentheses*{x, y} \in \brackets*{0, 63} \times \brackets*{0, 63}$ and selecting a frame per time interval of size $0.002$.
This discretization is used to solve \cref{eq.wave_eq} as its spatial derivatives are estimated thanks to finite differences; once computed, they are used in an ODE numerical solver to solve \cref{eq.wave_eq} on $t$.
Spatial derivatives are estimated with finite differences of order $5$, and the ODE solver is the fourth-order Runge-Kutta method with the $3/8$ rule \citep{Kutta1901, Hairer1993} and step size $0.001$.
The data are finally normalized following a min-max $\brackets*{0, 1}$ scaling per sequence.

The dataset is then split into training ($240$ sequences) and testing ($60$ sequences) sets.
Sequences sampled during training are random chuncks of length $\nu + 1 = 25$, including $\tau + 1 = 5$ conditioning frames, of full-size training sequences.
Sequences used during testing are all possible chunks of length $\tau + 1 + 40 = 45$ from full-size testing sequences.

Finally, WaveEq-100 is created from WaveEq by selecting $100$ pixels uniformly at random.
The extracted pixels are selected before training and are fixed for both training and testing.
Therefore, train and test sequences for WaveEq-100 consist of vector of size 100 extracted from WaveEq frames.
Training and testing sequences are chosen to be the same as those of WaveEq.

\subsection{Sea Surface Temperature}

SST is composed of sea surface temperatures of the Atlantic ocean generated using E.U. Copernicus Marine Service Information thanks to the state-of-the-art simulation engine NEMO.
The use of a so-called reanalysis procedure implies that these data accurately represent the actual temperature measures.
For more information, we refer to the complete description of the data by \cite{Bezenac2018}.
The data history of this engine is available online.\footnote{\url{https://resources.marine.copernicus.eu/?option=com_csw&view=details&product_id=GLOBAL_ANALYSIS_FORECAST_PHY_001_024}.} 
Unfortunately, due to recent maintenance, data history is limited to the last three years; prior histories should be manually requested.

The dataset uses daily temperature acquisitions from \formatdate{28}{12}{2006} to \formatdate{05}{04}{2017} of a $481 \times 781$ zone, from which $29$ zones of size $64 \times 64$ zones are extracted.
We follow the same setting as \cite{Bezenac2018} by training all models with $\tau + 1 = 4$ conditioning steps and $\nu - \tau = 6$ steps to predict, and evaluating them only on zones $17$ to $20$.
These zones are particularly interesting since they are the places where cold waters meet warm waters, inducing more pronounced motion.

We normalize the data in the same manner as \cite{Bezenac2018}.
Each daily acquisition of a zone is first normalized using the mean and standard deviation of measured temperatures in this zone computed for all days with the same date of the year from the available data (daily history climatological normalization).
Each zone is then normalized so that the mean and variance over all acquisitions correspond to those of a standard Gaussian distribution.
These normalized data are finally fed to the model; MSE scores reported in \cref{tab:phy-res} are computed once the performed normalization of the data and model prediction is reverted to the original temperature measurement space, in order to compute physically meaningful scores.

Training sequences correspond to randomly selected chunks of length $\nu = 10$ in the first \num{2987} acquisitions (corresponding to $80 \%$ of total acquisitions), and testing sequences to all possible chunks of length $\nu = 10$ in the remaining \num{747} acquisitions.

\subsection{Moving MNIST}

This dataset involves two MNIST digits \citep{LeCun1998} of size $28 \times 28$ that linearly move within $64 \times 64$ frames and deterministically bounce against frame borders following reflection laws.
We use the modified version of the dataset proposed by \cite{Franceschi2020} instead of the original one \citep{Srivastava2015}.
We train all models in the same setting as \cite{Denton2017}, with $\tau + 1 = 5$ conditioning frames and $\nu - \tau = 10$ frames to predict, and test them to predict either $10$ or $95$ frames ahead.
Training data consist in trajectories of digits from the MNIST training set, randomly generated on the fly during training.
Test data are produced by computing a trajectory for each digit of the MNIST testing set, and randomly pairwise combining them, thus producing \num{5000} sequences.

To evaluate disentanglement with content swapping, we report PSNR and SSIM metrics between the swapped sequence produced by our model and a ground truth.
However, having two digits in the image, there is an ambiguity as to in which order target digits should be swapped in the ground truth.
To account for this ambiguity and thanks to the synthetic nature of the dataset, we instead build two ground truth sequences for both possible digit swap permutations, and report the lowest metric between the generated sequence and both ground truths (i.e., we choose the closest ground truth to compare to with respect to the considered metric).

\subsection{3D Warehouse Chairs}

This multi-view dataset introduced by \cite{Aubry2014} contains \num{1393} three-dimensional models of chairs seen under the same periodic angles.
We resize the original $600 \times 600$ images by center-cropping them to $400 \times 400$ images, and downsample them to $64 \times 64$ frames using the Lanczos filter of the Pillow library.\footnote{\url{https://pillow.readthedocs.io/}}

We create sequences from this dataset for our model by assembling the views of each chair to simulate its rotation from right to left until it reaches its initial position.
This process is repeated for each existing angle to serve as initial position for all chairs.
We chose this dataset instead of \cite{Denton2017}'s multi-view chairs dataset because the latter contains too few objects to allow both tested methods to generalize on the testing set, preventing us to draw any conclusion from the experiment.
We train models on this dataset with $\tau + 1 = 5$ conditioning frames and $\nu - \tau = 10$ frames to predict, and test them to predict $15$ frames within the content swap experiment.
Training and testing data are constituted by randomly selecting $85 \%$ of the chairs for training and $15 \%$ of the remaining ones for testing.
Disentanglement metrics are computed similarly to the ones on Moving MNIST, but with only one reference ground truth corresponding to the chair given as content input at the position of the chair given as dynamic input.

\subsection{TaxiBJ}

This crowd flow dataset provided by \cite{Zhang2017} consists in two-channel $32 \times 32$ frames representing the inflow and outflow of taxis in Beijing, each pixel corresponding to a square region of the city.
Observations are registered every thirty minutes.
It is highly structured as the flows are dependent on the infrastructure of the city, and complex since methods have to account for non-local dependencies  and model subtle changes in the evolution of the flows.

We follow the preprocessing steps of \cite{Wang2018} and \cite{LeGuen2020} by performing a min-max normalization of the data to the $\brackets*{0, 1}$ range.
We train models on this dataset with $\tau + 1 = 4$ conditioning frames and $\nu - \tau = 4$ frames to predict, and test them to predict $4$ frames like our competitors on the last four weeks of data which are excluded from the training set.
MSE on this dataset is reported in the $\brackets*{0, 1}$-normalized space and multiplied by a hundred times the dimensionality of a frame, i.e. by $100 \times 32 \times 32 \times 2$.

\section{Training Details}
\label{sec:training}
Along with the code in the supplementary material, we provide in this section sufficient details in order to replicate our results.

\subsection{Reproduction of Baselines}

\paragraph{PKnl.}
We retrained PKnl \citep{Bezenac2018} on SST using the official implementation and the indicated hyperparameters.

\paragraph{SVG, MIM and DDPAE.}
We trained SVG \citep{Denton2018}, MIM \citep{Wang2019} and DDPAE \citep{Hsieh2018} on our version of Moving MNIST using the official implementation and the same hyperparameters that the authors used for the original version of Moving MNIST.

We trained MIM on SST using the recommended hyperparameters of the authors, and SVG by retaining the same hyperparameters as those used on KTH.

\paragraph{DrNet.}
We trained DrNet \citep{Denton2017} on our version of Moving MNIST using the same hyperparameters originally used for the alternative version of the dataset on which it was originally trained (with digits of different colors).
To this end, we reimplemented the official Lua implementation into a Python code in order to train it with a more recent infrastucture.
We also trained DrNet on 3D Warehouse Chairs using the same hyperparameters used by its authors on the smaller multi-view chairs dataset on which they trained their method.

\paragraph{PhyDNet.}
We trained PhyDNet \citep{LeGuen2020} on SST and our version of Moving MNIST using the official implementation and the same hyperparameters that the authors used for SST and the original version of Moving MNIST.
We removed the skip connections used by the authors on the Moving MNIST dataset in order to perform a fairer comparison with other models, such as ours, in our experimental study that do not incorporate skip connections on this dataset.

\subsection{Model Specifications}

\subsubsection{Implementation}

We used Python 3.8.1 and PyTorch 1.4.0 \citep{Paszke2019} to implement our model.
Each model was trained on an Nvidia GPU with CUDA 10.1.
Training is done with mixed-precision training \citep{Micikevicius2018} thanks to the Apex library.\footnote{\url{https://github.com/nvidia/apex}.}

\subsubsection{Architecture}

\paragraph{Combination of $S$ and $T$.}

As explained in \cref{sec:method}, the default choice of combination of $S$ and $T$ as decoder inputs is the concatenation of both vectorial variables: it is generic, and allows the decoder to learn an appropriate combination function $\zeta$ as in \cref{eq.explicit_var_sep}.

Nonetheless, further knowledge of the studied dataset can help to narrow the choices of combination functions.
Indeed, we choose to multiply $S$ and $T$ before giving them as input to the decoder for both datasets WaveEq and WaveEq-100, given the knowledge of the existence of product-separable solutions to the homogeneous version of equation (i.e., without source).
This shows that it is possible to change the combination function of $S$ and $T$, and that existing combination functions in the PDE literature could be leveraged for other datasets.

\paragraph{Encoders $E_{S}$ and $E_{T}$, and decoder $D$.}

For WaveEq, the encoder and decoder outputs are considered to be vectors; images are thus flattened before encoding and reshaped after decoding to $64 \times 64$ frames.
The encoder is a MultiLayer Perceptron (MLP) with two hidden layers of size $1200$ and internal ReLU activation functions.
The decoder is an MLP with three hidden layers of size $1200$, internal ReLU activation functions, and a final sigmoid activation function for the decoder.
The encoder and decoder used for WaveEq-100 are similar to those used for WaveEq, but with two hidden layers each, of respective sizes $2400$ and $150$.

We used for SST a VGG16 architecture \citep{Simonyan2015}, mirrored between the encoder and the decoder, complemented with skip connections integrated into $S$ \citep{Ronneberger2015} from all internal layers of the encoder to corresponding decoder layers, also leveraged by \cite{Bezenac2018} in their PKnl model.
We adapted this VGG16 architecture without skip connections for the $32 \times 32$ frames of TaxiBJ by removing the shallowest upsampling and downsampling operations in the VGG encoder and decoder.
For Moving MNIST, the encoder and its mirrored decoder are shaped with the DCGAN discriminator and generator architecture \citep{Radford2016}, with an additional sigmoid activation after the very last layer of the decoder; this encoder and decoder DCGAN architecture is also used by DrNet and DDPAE.
We highlight that we leveraged in both SST and Moving MNIST architectural choices that are also used in compared baselines, enabling fair comparisons.

For the two-dimensional latent space experiments on SST (see \cref{sst-model}), we use a modified version of the VGG encoder / decoder network by removing the two deepest maximum pooling layers, thus preserving the two-dimensional latent structures.
The decoder mirrors the encoder complemented with skip connections.

Regarding 3D Warehouse Chairs, we also followed the same architectural choices as DrNet with a ResNet18-like architecture for the encoders and a DCGAN architecture, followed by a sigmoid activation after the last layer for the decoder.

Encoders $E_{S}$ and $E_{T}$ taking as input multiple observations, we combine them by either concatenating them for the vectorial observations of WaveEq-100, or grouping them on the color channel dimensions for the other datasets where observations are frames.
Each encoder and decoder layer was initialized from a normal distribution with standard deviation $0.02$ (except for biases initialized to $0$, and batch normalizations weights drawn from a Gaussian distribution with unit mean and a standard deviation of $0.02$).

\paragraph{ODE solver.}
Following the recent line of work assimilating residual networks \citep{He2016} with ODE solvers \citep{Lu2018, Chen2018}, we use a residual network as an integrator for \cref{eq.time_ode}.
This residual network is composed of a given number $K$ of residual blocks, each block $i \in \lrbrackets*{1, K}$ implementing the application $\id + \texttt{$g_{i}$}$, where $g_{i}$ is an MLP with a two hidden layers of size $H$ and internal ReLU activation functions.
The parameter values for each dataset are:
\begin{itemize}
    \item WaveEq and WaveEq-100: $K = 3$ and $H = 512$;
    \item SST (with linear latent states): $K = 3$ and $H = 1024$;
    \item Moving MNIST, 3D Warehouse Chairs and TaxiBJ: $K = 1$ and $H = 512$.
\end{itemize}
Each MLP is orthogonally initialized with the following gain for each dataset:
\begin{itemize}
    \item WaveEq, WaveEq-100, SST (with linear latent states), 3D Warehouse Chairs and TaxiBJ: $0.71$;
    \item Moving MNIST: $1.41$.
\end{itemize}

For SST with two-dimensional states, the MLPs are replaced by convolutional layers with kernel size $3$, padding $1$ and a number of hidden channels equal to $H = 128$.
We set $K = 2$ and an orthogonal initialization gain of $0.2$.
ReLU activations are replaced by Leaky ReLU activations and preceded by batch normalization layers.

\paragraph{Latent variable sizes.}
$S$ and $T$ have the following vectorial dimensions for each dataset:
\begin{itemize}
    \item WaveEq and WaveEq-100: $32$;
    \item SST, respectively $196\times16\times16$ and $64\times16\times16$; for the linear version, both are set to $256$.
    \item Moving MNIST and TaxiBJ: respectively, $128$ and $20$;
    \item 3D Warehouse Chairs: respectively, $128$ and $10$.
\end{itemize}

Note that, in order to perform fair comparisons, the size of $T$ for baselines without static component $S$ is chosen to be the sum of the vectorial sizes of $S$ and $T$ in the full model.
The skip connections of $S$ for SST cannot, however, be integrated into $T$, as its evolution is only modeled in the latent space, and it is out of the scope of this paper to leverage low-level dynamics.

\subsection{Optimization}

Optimization is performed using the Adam optimizer \citep{Kingma2015} with initial learning rate \num{4e-4} for WaveEq, WaveEq-100, Moving MNIST, 3D Warehouse Chairs and SST and \num{4e-5} for TaxiBJ, and with decay rates $\beta_{1} = 0.9$ (except for the experiments on Moving MNIST where we choose $\beta_{1} = 0.5$) and $\beta_{2} = 0.99$.

\paragraph{Loss function.}
Chosen coefficients values of $\lambda_{\mathrm{pred}}$, $\lambda_{\mathrm{AE}}$, $\lambda_{\mathrm{reg}}^{S}$, and $\lambda_{\mathrm{reg}}^{T}$ are the following:
\begin{itemize}
    \item $\lambda_{\mathrm{pred}} = 45$;
    \item $\lambda_{\mathrm{AE}} = 45$ for TaxiBJ; $10$ for SST (linear)and Moving MNIST; $1$ for WaveEq, WaveEq-100 and 3D Warehouse Chairs; $0.1$
    for SST;
    \item $\lambda_{\mathrm{reg}}^{S} = 100$ for SST; $\lambda_{\mathrm{reg}}^{S} = 45$ for WaveEq, WaveEq-100, SST (linear) and Moving MNIST; $1$ for 3D Warehouse Chairs; $0.0001$ for TaxiBJ;
    \item $\lambda_{\mathrm{reg}}^{T} = \frac{1}{2} p \times 10^{-3}$ for WaveEq, WaveEq-100, Moving MNIST, 3D Warehouse Chairs and TaxiBJ (where $p$ is the dimension of $T$); $\frac{1}{2} p \times 10^{-2}$ for SST (linear); $5 \times 10^{-6}$ for SST.
\end{itemize}

The batch size is chosen to be $128$ for WaveEq, WaveEq-100, Moving MNIST and 3D Warehouse Chairs, and $100$ for SST and TaxiBJ.

\paragraph{Training length.}
The number of training epochs for each dataset is:
\begin{itemize}
    \item WaveEq and WaveEq-100: $250$ epochs;
    \item SST: $30$ epochs; SST (linear): $80$ epochs;
    \item Moving MNIST: $800$ epochs, with an epoch corresponding to \num{200000} trajectories (the dataset being infinite), and with the learning rate successively divided by $2$ at epochs $300$, $400$, $500$, $600$, and $700$;
    \item 3D Warehouse Chairs: $120$ epochs;
    \item TaxiBJ: $550$ epochs, with the learning rate divided by $5$ at epochs $250$, $300$, $350$, $400$ and $450$.
\end{itemize}

\subsection{Prediction Offset for SST}

Using the formalism of our work, our algorithm trains to reconstruct $v = \parentheses*{v_{t_{0}}, \ldots, v_{t_{1}}}$ from conditioning frames $V_\tau\parentheses*{t_{0}}$.
Therefore, it first learns to reconstruct $V_\tau\parentheses*{t_{0}}$.

However, the evolution of SST data is chaotic and predicting above an horizon of $6$ with coherent and sharp estimations is challenging.
Therefore, for the SST dataset only, we chose to supervise the prediction from $t = t_{0} + \parentheses*{\tau+1} \Delta t$, i.e, our algorithm trains to forecast $v_{t_{0} + \parentheses*{\tau + 1} \Delta t}, \ldots, v_{t_{1}}$ from $V_\tau\parentheses*{t_{0}}$.
It simply consists in making the temporal representation $E_T \parentheses*{V_\tau\parentheses*{t_{0}}}$ match the observation $v_{t_{0} + \parentheses*{\tau + 1} \Delta t}$ instead of $v_{t_{0}}$.
This index offset does not change our interpretation of spatiotemporal disentanglement through separation of variables.

\section{Additional Results and Samples}
\label{sec:additional}

\subsection{Ablation Study on Moving MNIST}

\begin{table}
    \caption{
        \label{tab:moving-mnist-ablation}
        Prediction and content swap PSNR and SSIM scores of variants of our model.
    }
    \sisetup{detect-weight, table-align-uncertainty=true, table-number-alignment=center, mode=text}
    \renewrobustcmd{\bfseries}{\fontseries{b}\selectfont}
    \renewrobustcmd{\boldmath}{}
    \centering
    \vspace{\baselineskip}
    \begin{tabular}{lS[table-format=2.2]S[table-format=1.4]S[table-format=2.2]S[table-format=1.4]S[table-format=2.2]S[table-format=1.4]S[table-format=2.2]S[table-format=1.4]}
        \toprule
        \multirow{2}{*}[-0.5\dimexpr \aboverulesep + \belowrulesep + \cmidrulewidth]{Models} & \multicolumn{2}{c}{Pred. ($t + 10$)} & \multicolumn{2}{c}{Pred. ($t + 95$)} & \multicolumn{2}{c}{Swap ($t + 10$)} & \multicolumn{2}{c}{Swap ($t + 95$)} \\
        \cmidrule(lr){2-3} \cmidrule(lr){4-5} \cmidrule(lr){6-7} \cmidrule(lr){8-9}
        & {PSNR} & {SSIM} & {PSNR} & {SSIM} & {PSNR} & {SSIM} & {PSNR} & SSIM \\
        \midrule
        Ours & \bfseries 21.70 & \bfseries 0.9088 & \bfseries 17.50 & \bfseries 0.7990 & \bfseries 18.42 & \bfseries 0.8368 & \bfseries 16.50 & \bfseries 0.7713 \\
        Ours (without $S$) & 20.46 & 0.8867 & 14.95 & 0.6707 & {\textemdash} & {\textemdash} & {\textemdash} & {\textemdash} \\
        Ours ($\lambda_{\mathrm{AE}} = 0$) & 21.61 & 0.9058 & 16.58 & 0.7611 & 18.21 & 0.8309 & 15.79 & 0.7399 \\
        Ours ($\lambda_{\mathrm{reg}}^{S} = 0$) & 15.99 & 0.6900 & 12.31 & 0.5702 & 13.73 & 0.5476 & 12.07 & 0.5556 \\
        Ours ($\lambda_{\mathrm{reg}}^{T} = 0$) & 15.63 & 0.7369 & 14.02 & 0.7253 & 14.91 & 0.7154 & 13.95 & 0.7234 \\
        Ours (GRU) & 21.66 & 0.9088 & 15.45 & 0.4888 & 17.70 & 0.8178 & 14.77 & 0.4718 \\
        \bottomrule
    \end{tabular}
\end{table}

We report in \cref{tab:moving-mnist-ablation} the results of an ablation study of our model on Moving MNIST, that we comment in \cref{sec:mnist-exp}.

\subsection{Preliminary Results on KTH}

The application of our method to natural videos is an interesting perspective, but would motivate further adaptation of the model (see perspectives in the conclusion), in particular regarding the integration of stochastic dynamics.
Indeed, there is a consensus in the literature (e.g.: \cite{Denton2018, Villegas2019, Weissenborn2020}) indicating that human motion datasets require stochastic modeling because of the inherently highly random events occurring in these videos.
Tackling this issue would require to incorporate stochasticity in our model, for example leveraging variational autoencoders like \cite{Denton2018}, or supplement it with adversarial losses on the image space, for instance like \cite{Mathieu2016} and \cite{Lee2018}.
These changes are feasible, but are out of the scope of this paper.

\begin{table}
    \caption{
        \label{tab:kth}
        FVD score of compared models on KTH. The bold score indicates the best performing method.
    }
    \sisetup{detect-weight, table-align-uncertainty=true, table-number-alignment=center, mode=text}
    \renewrobustcmd{\bfseries}{\fontseries{b}\selectfont}
    \renewrobustcmd{\boldmath}{}
    \centering
    \vspace{\baselineskip}
    \begin{tabular}{S[table-format=3]S[table-format=3]S[table-format=3]S[table-format=3]}
        \toprule
        {Ours} & {PhyDNet} & {SVG} & {DrNet} \\
        \midrule
        \bfseries 330 & 384 & 375 & 383 \\
        \bottomrule
    \end{tabular}
\end{table}

Nonetheless, we investigate the realistic video dataset KTH \citep{Schuldt2004}, which is an action recognition video database featuring various subjects performing actions in front of different backgrounds.
We trained our model, SVG, DrNet and PhyDNet on this dataset.
DrNet and PhyDNet are powerful deterministic approaches, while SVG is a standard stochastic video prediction model.
We compare all models in terms of FVD \citep[lower is better]{Unterthiner2018}, which is a metric based on deep features that evaluates the realism of the generated videos.

Results are reported in \cref{tab:kth}.
We observe that our model substantially outperforms the considered baselines.
These significant results against powerful deterministic baselines, and even the standard stochastic method SVG, confirm our advantage at modeling complex dynamics and support our claim that our model lays the foundations for domain-specific methods, such as a stochastic version for natural videos.

\paragraph{Reproductibility.}

We use the following training parameters for KTH:
\begin{itemize}
    \item we follow the same dataset processing and evaluation procedure as \cite{Denton2018};
    \item we train our model on $125$ epochs with batch size $100$, with an epoch being defined as \num{100000} training sequences;
    \item we set the learning rate to $2 \times 10^{-4}$ and the same optimizer parameters as for SST;
    \item $\lambda_{\mathrm{pred}} = 45$, $\lambda_{\mathrm{AE}} = 10 = \lambda_{\mathrm{reg}}^{S} = 10$, $\lambda_{\mathrm{reg}}^{T} = p \times 10^{-4}$;
    \item the size of $S$ and $T$ are respectively $128$ and $50$;
    \item the ODE is solved with a flat latent architecture and parameters $K = 1$ and $H = 512$;
    \item the encoder and decoder architecture is VGG16 with skip connections integrated into $S$ from $E_{S}$ to $D$, and with the decoder output being given to a final sigmoid activation.
\end{itemize}

We reproduced SVG, DrNet and PhyDNet using the recommended hyperparameters of their authors.
We trained PhyDNet for $125$ epochs, like our model, to obtain a fair evaluation despite its low efficiency (six times slower than ours).

\subsection{Modeling SST with Separation of Variables}
\label{sst-model}

\begin{table}
    \caption{
        \label{tab:phy-res-appendix}
        Forecasting performance on SST of PKnl, PhyDNet and our model with respect to indicated prediction horizons.
        Bold scores indicate the best performing method.
    }
    \sisetup{detect-weight, table-align-uncertainty=true, table-number-alignment=center, mode=text}
    \renewrobustcmd{\bfseries}{\fontseries{b}\selectfont}
    \renewrobustcmd{\boldmath}{}
    \centering
    \vspace{\baselineskip}
    \begin{tabular}{lS[table-format=1.2]S[table-format=1.2]S[table-format=1.4]S[table-format=1.4]}
        \toprule
        \multirow{2}{*}[-0.5\dimexpr \aboverulesep + \belowrulesep + \cmidrulewidth]{Models} & \multicolumn{2}{c}{MSE} & \multicolumn{2}{c}{SSIM} \\
        \cmidrule(lr){2-3} \cmidrule(lr){4-5}
        & {$t + 6$} & {$t + 10$} & {$t + 6$} & {$t + 10$} \\
        \midrule
        PKnl & 1.28 & 2.03 &  0.6686 & 0.5844 \\
        PhyDNet & 1.27 & 1.91 & 0.5782 & 0.4645 \\
        SVG & 1.51 & 2.06 & 0.6259 & 0.5595 \\
        MIM & 0.91 & 1.45 & 0.7406 & 0.6525 \\
        \midrule
        Ours & \bfseries 0.86 & \bfseries 1.43 & \bfseries 0.7466 & \bfseries 0.6577 \\
        Ours (without $S$) & 0.95 & 1.50 & 0.7204 & 0.6446 \\
        \midrule
        Ours (linear) & 1.15 & 1.80 & 0.6837 & 0.5984 \\
        Ours (linear, without $S$) & 1.46 & 2.19 & 0.6200 & 0.5456 \\
        \bottomrule
    \end{tabular}
\end{table}

We present in \cref{tab:phy-res-appendix} results of \cref{tab:phy-res} for SST, complemented with an alternative version of our model obtained using vectorial representation for $S$ and $T$ and MLPs to compute the derivative of $T$.
The latter setting corresponds to a strictly enforced separation of spatial and dynamical variables, with results significantly outperforming powerful methods PhyDNet, PKnl and SVG thanks to this separation, as attested by the corresponding ablation without a static component.

However, sea surface temperature exhibits highly local structure that can be assimilated to a flow in a coarse approximation.
For example, there is transport of large bodies of hot and cold water.
Accordingly, performances may be enhanced by considering local dependencies in the dynamics, as also implemented by MIM and PhyDNet.
We propose to do so by considering like the latter methods two-dimensional latent states for the static $S$ and the dynamical $T$, and convolutional networks to model the derivative of $T$.

Accounting for such locality in the dynamics amounts to implementing another separation than the usual separation between $t$ and spatial variables.
Indeed, it rather excludes unknown content variables from the dynamics.
The resulting dynamics is then a PDE over time $t$ and the observation coordinates $x$ and $y$ that we implement using convolutional neural networks, following \cite{Long2018} and \cite{Ayed2020}.
This different kind of separation of variables simplifies learning by estimating a PDE that is simpler than the original one, since it acts on fewer variables.
It highlights the generality of our intuition of using the separation of variables, which may be used in other settings that strict spatiotemporal disentanglement.
This approach, while still maintaining disentangling properties, significantly improves prediction performances.

Note that our proposition remains computationally much lighter than the alternatives MIM, PhyDNet and SVG.

\subsection{Additional Samples}

\subsubsection{WaveEq}

\begin{figure}
    \centering
    \includegraphics[width=\textwidth, trim={0 5.3cm 0 3.5cm}, clip]{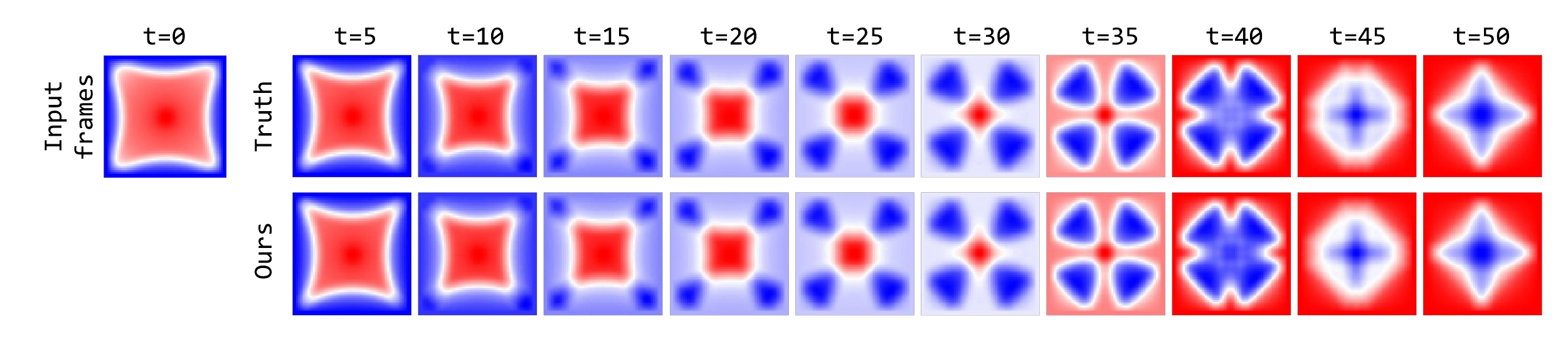}
    \caption{\label{fig:wave-eq}Example of predictions of our model on WaveEq.}
\end{figure}

\begin{figure}
    \centering
    \includegraphics[width=\textwidth, trim={2cm 7cm 1cm 4cm}, clip]{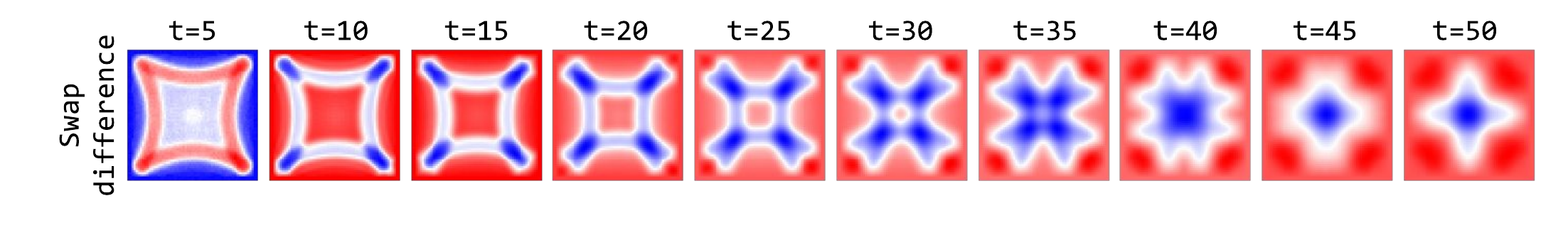}
    \caption{\label{fig:wave-swap}Evolution of the scaled difference between the forecast of a sequence and the same forecast with a spatial code coming from another sequence for the WaveEq dataset.}
\end{figure}

We provide in \cref{fig:wave-eq} a sample for the WaveEq dataset, highlighting the long-term consistency in the forecasts of our algorithm.

We also show in \cref{fig:wave-swap} the effect in forecasting of changing the spatial code $S$ from the one of another sequence.

\subsubsection{SST}

\begin{figure}
    \centering
    \includegraphics[width=0.6\textwidth]{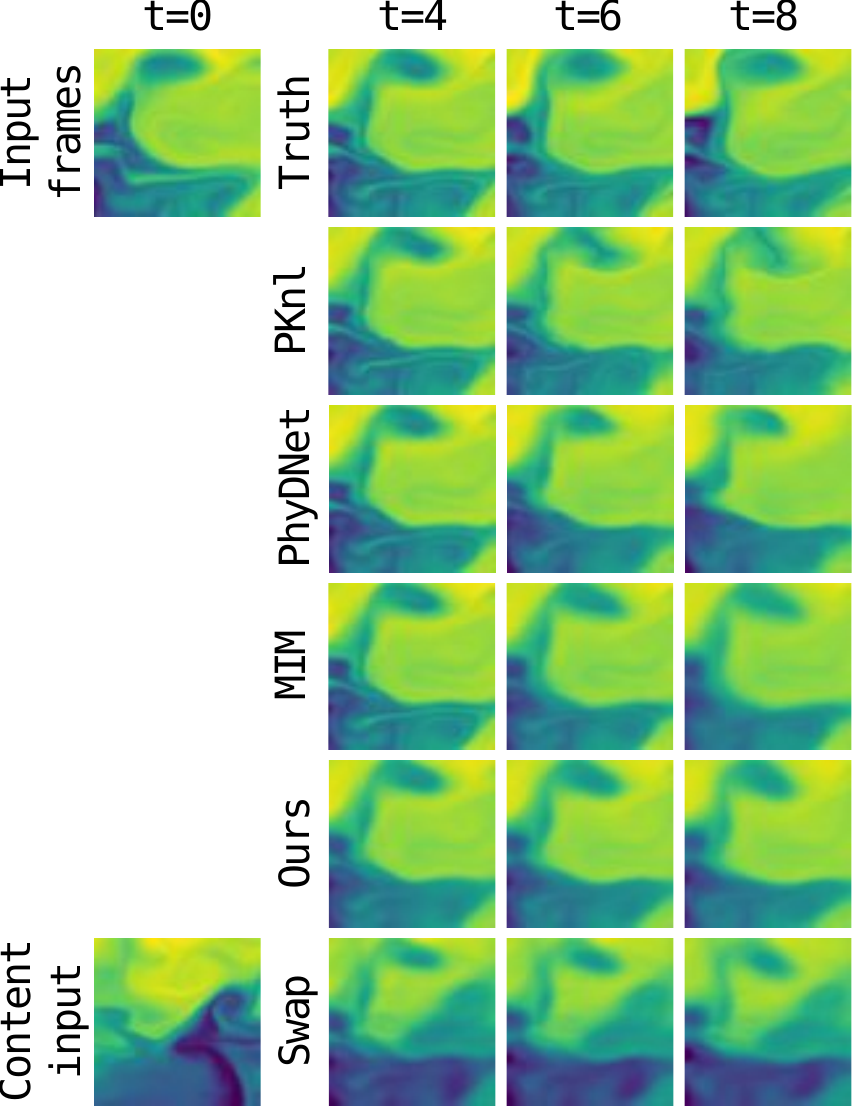}
    \caption{\label{fig:sst-supp}Example of predictions of compared models on SST.}
\end{figure}

We provide an additional sample for SST in \cref{fig:sst-supp}.

\subsubsection{Moving MNIST}

\begin{figure}
    \centering
    \includegraphics[width=\textwidth]{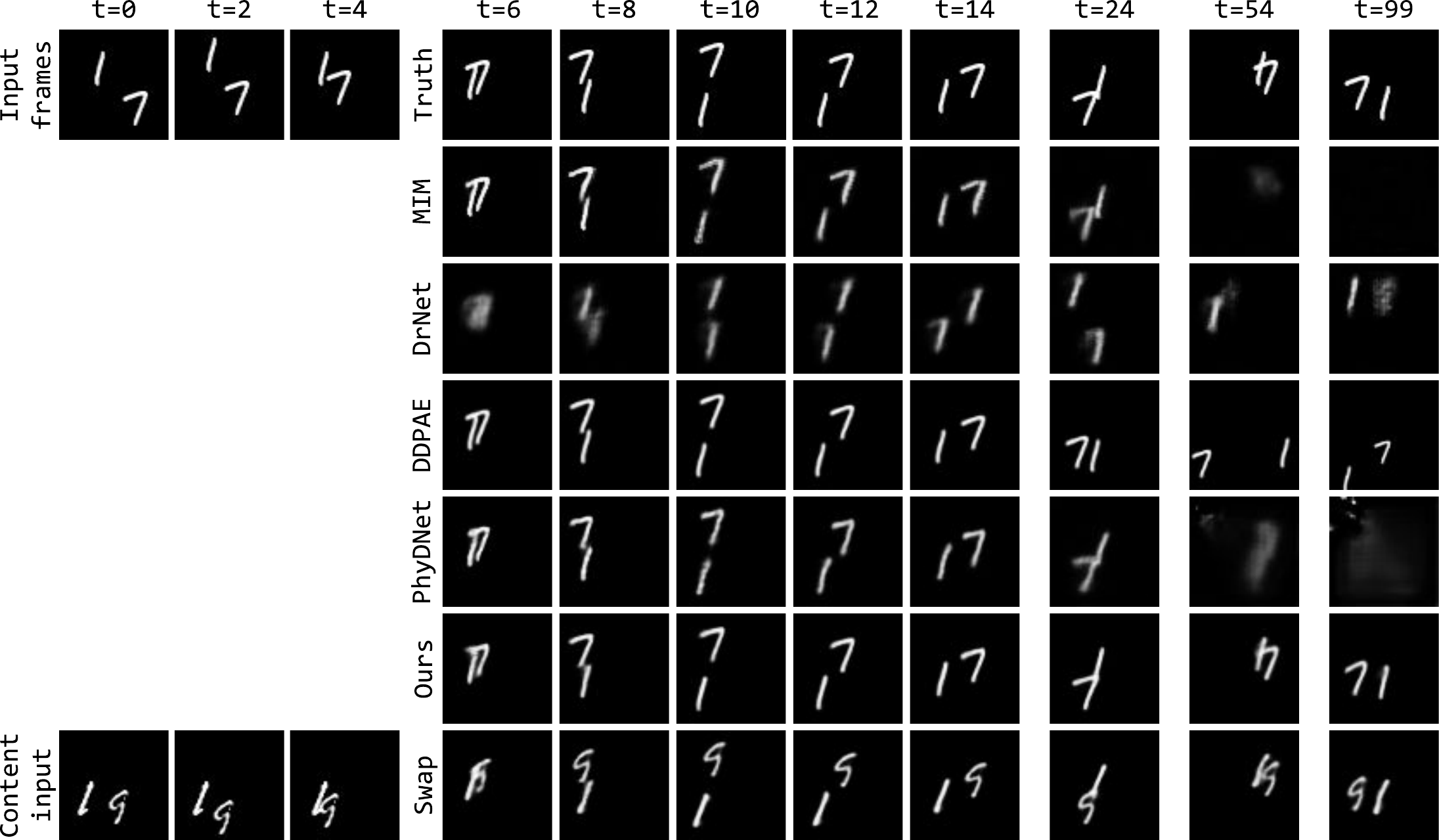}
    \caption{\label{fig:mnist-supp-1}Example of predictions of compared models on Moving MNIST.}
\end{figure}

\begin{figure}
    \centering
    \includegraphics[width=\textwidth]{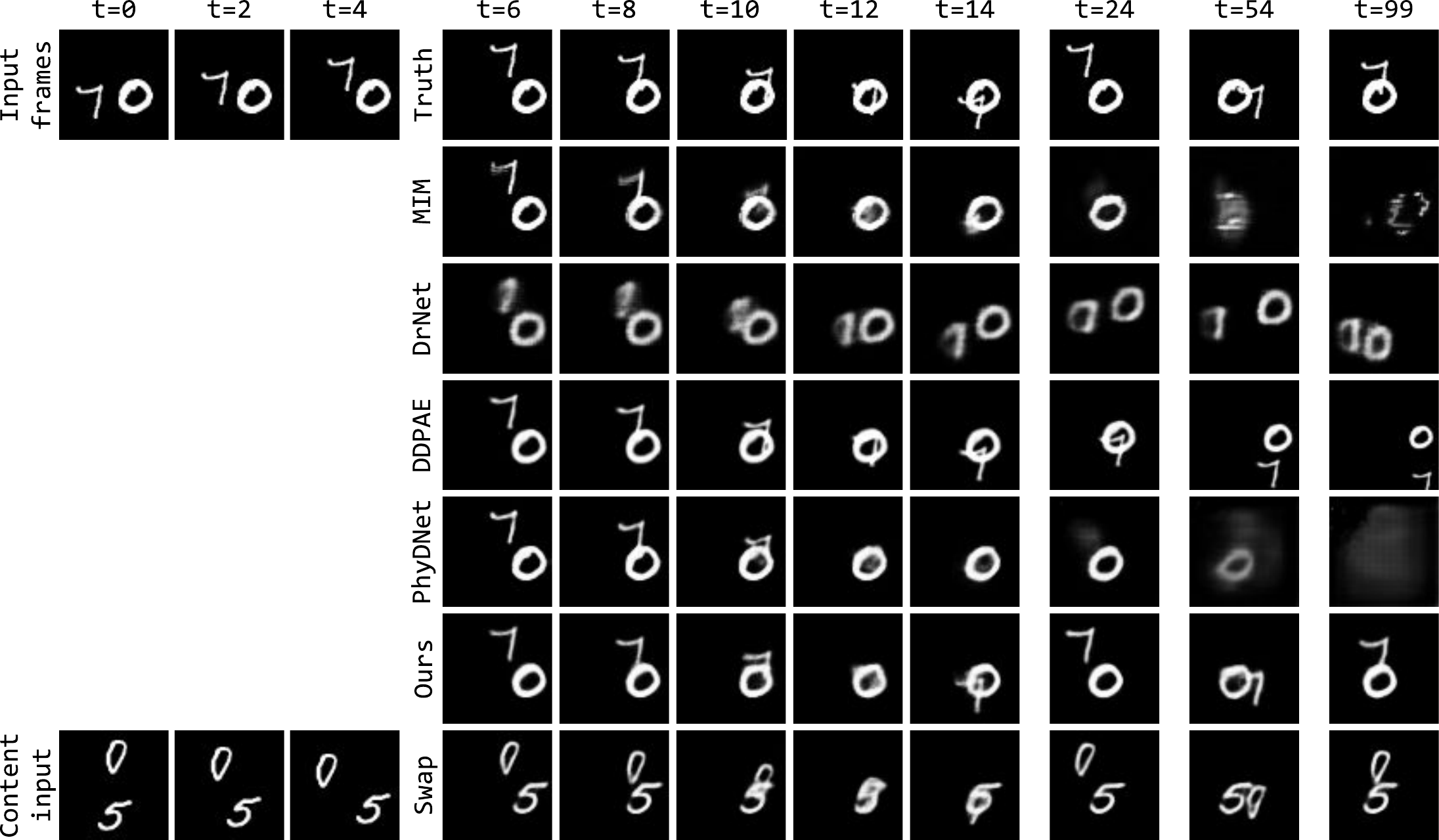}
    \caption{\label{fig:mnist-supp-2}Example of predictions of compared models on Moving MNIST.}
\end{figure}

We provide two additional samples for Moving MNIST in \cref{fig:mnist-supp-1,fig:mnist-supp-2}.

\subsubsection{3D Warehouse Chairs}

\begin{figure*}
    \subfigure[DrNet.]{\includegraphics[width=0.48\textwidth]{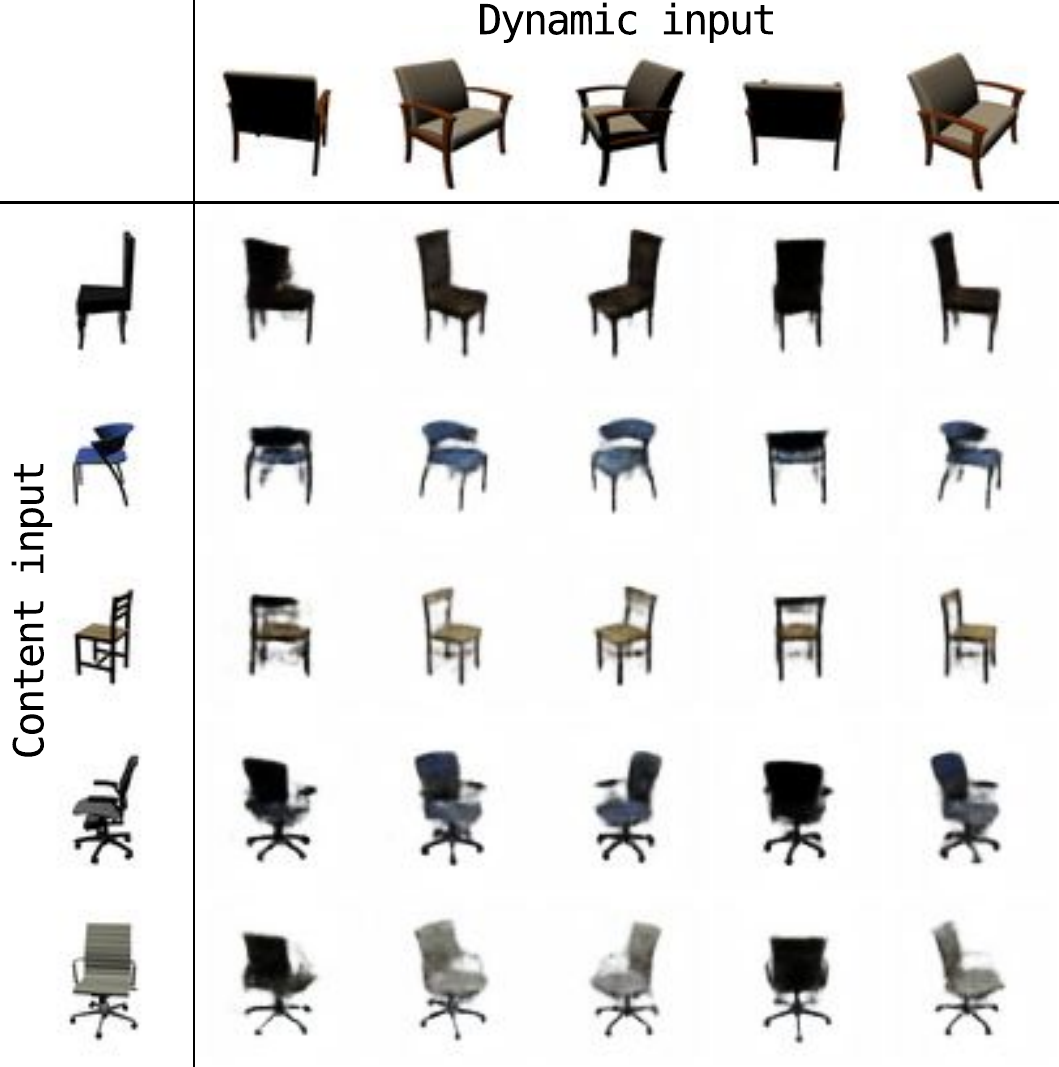}}
    \hfill
    \subfigure[Ours.]{\includegraphics[width=0.48\textwidth]{fig/chairs.pdf}}
    \caption{\label{fig:ChairsOursDenton}Fusion of content (first column) and dynamic (first row) variables in DrNet and our model on 3D Warehouse Chairs.}
\end{figure*}

We provide a qualitative comparison for the content swap experiment between our model and DrNet for 3D Warehouse Chairs in \cref{fig:ChairsOursDenton}.
We notice that DrNet produces substantially more blurry samples than our model and has difficulties to capture the exact dynamic of the chairs.

\end{document}